\documentclass{article}

\usepackage{microtype}
\usepackage{graphicx}
\usepackage{subcaption}
\usepackage{booktabs} % for professional tables
\usepackage[table]{xcolor} % for colored tables

\usepackage{hyperref}

\usepackage[preprint]{icml2026}

\usepackage{amsmath}
\usepackage{amssymb}
\usepackage{mathtools}
\usepackage{amsthm}
\usepackage{enumitem}

\usepackage[capitalize,noabbrev]{cleveref}

\theoremstyle{plain}

\theoremstyle{definition}

\theoremstyle{remark}

\usepackage[disable,textsize=tiny]{todonotes}
\usepackage{tcolorbox}
\newtcolorbox{promptbox}[1][]{
  colback=gray!5,
  colframe=gray!50,
  boxrule=0.5pt,
  arc=2pt,
  left=12pt, right=12pt, top=10pt, bottom=10pt,
  fontupper=\normalsize\ttfamily,
  title={},
  #1
}

\newtcolorbox{summarybox}[1][]{
  colback=blue!5,
  colframe=blue!50,
  boxrule=0.5pt,
  arc=3pt,
  left=5pt, right=5pt, top=5pt, bottom=5pt,
  fontupper=\small,
  title={#1},
  fonttitle=\bfseries\small
}

\icmltitlerunning{Thinking Out of Order: Order Robustness in Diffusion LMs}

\begin{document}

\twocolumn[
  \icmltitle{Thinking Out of Order: When Output Order Stops Reflecting Reasoning Order in Diffusion Language Models}



  \icmlsetsymbol{equal}{*}
  \icmlsetsymbol{advising}{$\dagger$}

  \begin{icmlauthorlist}
    \icmlauthor{Longxuan Yu}{equal,ucr}
    \icmlauthor{Yu Fu}{equal,ucr}
    \icmlauthor{Shaorong Zhang}{ucr}
    \icmlauthor{Hui Liu}{independent}
    \icmlauthor{Mukund Varma T}{ucsd}
    \icmlauthor{Greg Ver Steeg}{advising,ucr}
    \icmlauthor{Yue Dong}{advising,ucr}
  \end{icmlauthorlist}

  \icmlaffiliation{ucr}{University of California, Riverside, CA, USA}
  \icmlaffiliation{independent}{Independent Researcher}
  \icmlaffiliation{ucsd}{University of California, San Diego, CA, USA}

  \icmlcorrespondingauthor{Yue Dong}{yuedongle@gmail.com}

  \icmlkeywords{Diffusion Language Models, Order Robustness, Chain-of-Thought Reasoning, Masked Diffusion}

  \vskip 0.3in
]



\printAffiliationsAndNotice{\icmlEqualContribution \quad $\dagger$Equal advising.}

\begin{abstract}

    % Instruction following is a core capability of autoregressive (AR) models, developed through instruction tuning to faithfully adhere to user instructions. However, this strength becomes a liability when combined with autoregressive generation: if the prompt explicitly requests an output order that conflicts with natural reasoning order (e.g., ``answer first, then explain''), AR models are forced to produce answers before completing the underlying reasoning, leading to degraded accuracy.
    % Unlike AR models, diffusion language models generate all tokens in a block simultaneously, potentially avoiding this constraint. 
  
Autoregressive (AR) language models enforce a fixed left-to-right generation order, creating a fundamental limitation when the required output structure conflicts with natural reasoning (e.g., producing answers before explanations due to presentation or schema constraints). In such cases, AR models must commit to answers before generating intermediate reasoning, and this rigid constraint forces premature commitment. Masked diffusion language models (MDLMs), which iteratively refine all tokens in parallel, offer a way to decouple computation order from output structure.
We validate this capability on GSM8K, Math500, and ReasonOrderQA, a benchmark we introduce with controlled difficulty and order-level evaluation. When prompts request answers before reasoning, AR models exhibit large accuracy gaps compared to standard chain-of-thought ordering (up to 67\% relative drop), while MDLMs remain stable ($\leq$14\% relative drop), a property we term \emph{order robustness}.
Using ReasonOrderQA, we present evidence that MDLMs achieve order robustness by stabilizing simpler tokens (e.g., reasoning steps) earlier in the diffusion process than complex ones (e.g., final answers), enabling reasoning tokens to stabilize before answer commitment. Finally, we identify failure conditions where this advantage weakens, outlining the limits required for order robustness.

    % Diffusion language models offer an alternative generation paradigm by producing all tokens simultaneously through iterative refinement. We hypothesize that this parallel generation can avoid the output order constraint faced by AR models. 
    % Our experiments on GSM8K confirm this hypothesis: when prompts request answers before reasoning, AR models suffer severe accuracy drop ($\sim$67\%), while diffusion models remain stable ($<$4\%), a property we term \emph{order robustness}. 
    % We attribute this to token complexity: simpler tokens (e.g., reasoning steps) stabilize earlier than complex ones (e.g., final answers), enabling internal reasoning before answer commitment.
    % To systematically investigate this mechanism and its boundaries, we introduce ReasonOrderQA, a controlled benchmark with varying token complexity levels. Our analysis identifies two breakdown conditions: order robustness diminishes when (1) tokens have similar complexity, making it impossible to distinguish decoding priority, or (2) generation length is large, flattening the confidence landscape.

\end{abstract}

\section{Introduction}

\begin{figure}[t]
    \centering
    \includegraphics[width=\columnwidth]{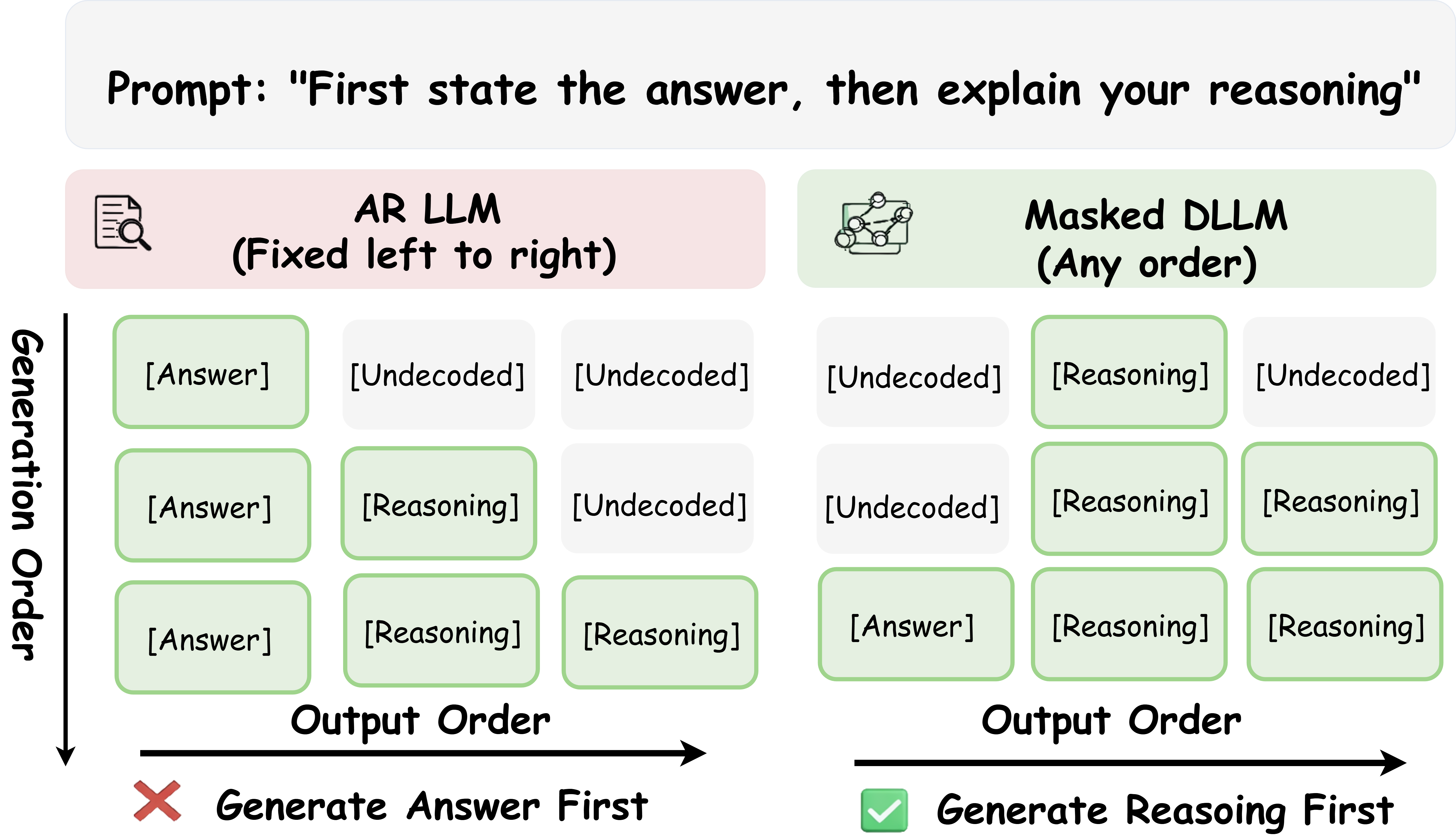}
    \caption{
       \textbf{Masked Diffusion models reason before answering—even when asked to answer first.}
       Under ``answer-first'' prompting, AR models must commit to answers before reasoning (generation order = output order). Masked diffusion LMs decode reasoning tokens first regardless of output position, enabling \emph{order robustness}.
    }
    \label{fig:teaser}
\end{figure}

Instruction-following is a fundamental capability of modern language models, enabling them to adhere to user-specified constraints on output format and structure~\citep{ouyang2022training, wei2021finetuned}. Chain-of-thought (CoT) prompting~\citep{kojima2022large, wei2022chain} extends this capability to complex reasoning tasks by instructing models to generate intermediate reasoning steps before producing final answers, significantly improving performance across diverse domains. However, this success depends on a critical assumption: the output order aligns with the model's internal reasoning order. In practice, this assumption is often violated by deployment constraints that prioritize application needs, such as the Minto Pyramid principle~\citep{minto2009pyramid} mandating conclusions before arguments to optimize human comprehension, or JSON schemas~\citep{tam2024let} enforcing fixed key sequences that force models to produce complex outputs (e.g., summary statistics) before the data needed to derive them.

For autoregressive (AR) language models, this mismatch exposes a fundamental limitation of left-to-right generation. AR models factorize sequences strictly in temporal order~\citep{vaswani2017attention, brown2020language}, coupling the \emph{generation order} (the sequence of probability assignments) with the \emph{output order} (the sequence of emitted tokens). When answers must precede reasoning in the output, AR models are forced to commit to answer tokens before the intermediate reasoning tokens exist in the generated context. This premature commitment constrains the model’s ability to revise early predictions and has been shown to limit non-monotonic generation and bidirectional planning~\citep{welleck2019nonmonotonic, gu2019indigo}. Yet its consequences for instruction-constrained reasoning remain largely unexplored.

Masked diffusion language models (MDLMs) offer an alternative generation paradigm. Rather than producing tokens sequentially, MDLMs iteratively refine a full-length sequence using a bidirectional Transformer and a sampling procedure that determines which positions to finalize at each step~\citep{nie2025llada, sahoo2024simple}. This process decouples computation order from textual position, allowing token estimates to evolve based on global sequence context and potentially enabling internal reasoning to emerge before answer commitment, even when answers appear first in the output. While recent work has primarily evaluated diffusion LMs in terms of final output quality and inference efficiency~\citep{wu2025fast, wang2025diffusion}, their ability to support reliable reasoning under strict output-order constraints has not been systematically examined.

To investigate this capability, we study \emph{order robustness}, defined as the ability to maintain reasoning accuracy when the prompt-specified output order is reversed (CoT-First versus Answer-First). Across GSM8K, Math500, and a new benchmark we introduce, we uncover a striking contrast between generation paradigms. Under Answer-First prompting, Qwen~\citep{qwen2.5} (autoregressive) exhibits large relative accuracy drops (up to $\sim$67\%), whereas LLaDA~\citep{nie2025llada} (diffusion trained from scratch) remains largely stable with only a 4\% relative drop. Dream~\citep{ye2025dream}, a diffusion model distilled from autoregressive weights, shows intermediate degradation ($\sim$46\%), indicating that order robustness depends not only on architecture but also on training methodology.

To make the underlying mechanism measurable rather than anecdotal, we introduce \textbf{ReasonOrderQA}, a controlled benchmark with graded arithmetic complexity (D1-D4) and explicit order-level evaluation. As illustrated in Figure~\ref{fig:teaser}, we observe a consistent \emph{complexity-driven stabilization} pattern in diffusion generation: lower-complexity reasoning tokens achieve high confidence and stabilize earlier than higher-complexity answer tokens, enabling internal reasoning to be refined before answer commitment regardless of output position. We further probe the boundaries of this mechanism and identify specific conditions under which order robustness breaks down.

Our contributions:
\setlist{nolistsep}
\begin{itemize}[noitemsep]
    \item We introduce \textbf{ReasonOrderQA}, a controlled benchmark with graded difficulty and order-level evaluation, enabling fine-grained analysis of order sensitivity across different token complexity levels.
    \item Our detailed experiments provide evidence for \textbf{order-robust reasoning} in masked diffusion LMs: unlike AR models that degrade significantly under answer-first prompting, diffusion models maintain accuracy regardless of output order.
    \item We systematically identify the mechanism for such order-robustness i.e. diffusion sampling automatically delays commitment on low-confidence tokens such as the final answers while high confidence tokens that convey reasoning stabilize earlier.
    \item Finally, we identify failure cases where order robustness weakens, highlighting the limits of this implicit effect and identifying directions for reliable order-agnostic generation.

\end{itemize}

% \noindent In the ongoing debate between autoregressive and diffusion language models, we identify an instance when diffusion-style generation may be preferable: when the required output order conflicts with the model's natural reasoning order
% \noindent Together, these findings suggest that prompt-specified output format need not dictate token stabilization order--offering a path toward models that can comply with diverse format requirements while preserving reasoning quality.

\section{Preliminary}
\label{sec:background}

We define \textbf{order robustness} as the ability to maintain reasoning accuracy when the prompt-specified output order is altered (e.g., CoT-First vs. Answer-First).

\paragraph{Autoregressive Models.}
Autoregressive (AR) models factorize the joint probability of a sequence $\mathbf{x} = [x_1, \dots, x_L]$ as:
\begin{equation}
    p_{\text{AR}}(\mathbf{x}) = \prod_{i=1}^{L} p(x_i \mid x_{<i})
\end{equation}
This enforces a strict left-to-right dependency: the prediction of token $x_i$ can only attend to past context $x_{<i}$, coupling generation order with textual position.

\begin{figure*}[t]
    \centering
    \includegraphics[width=0.8\linewidth]{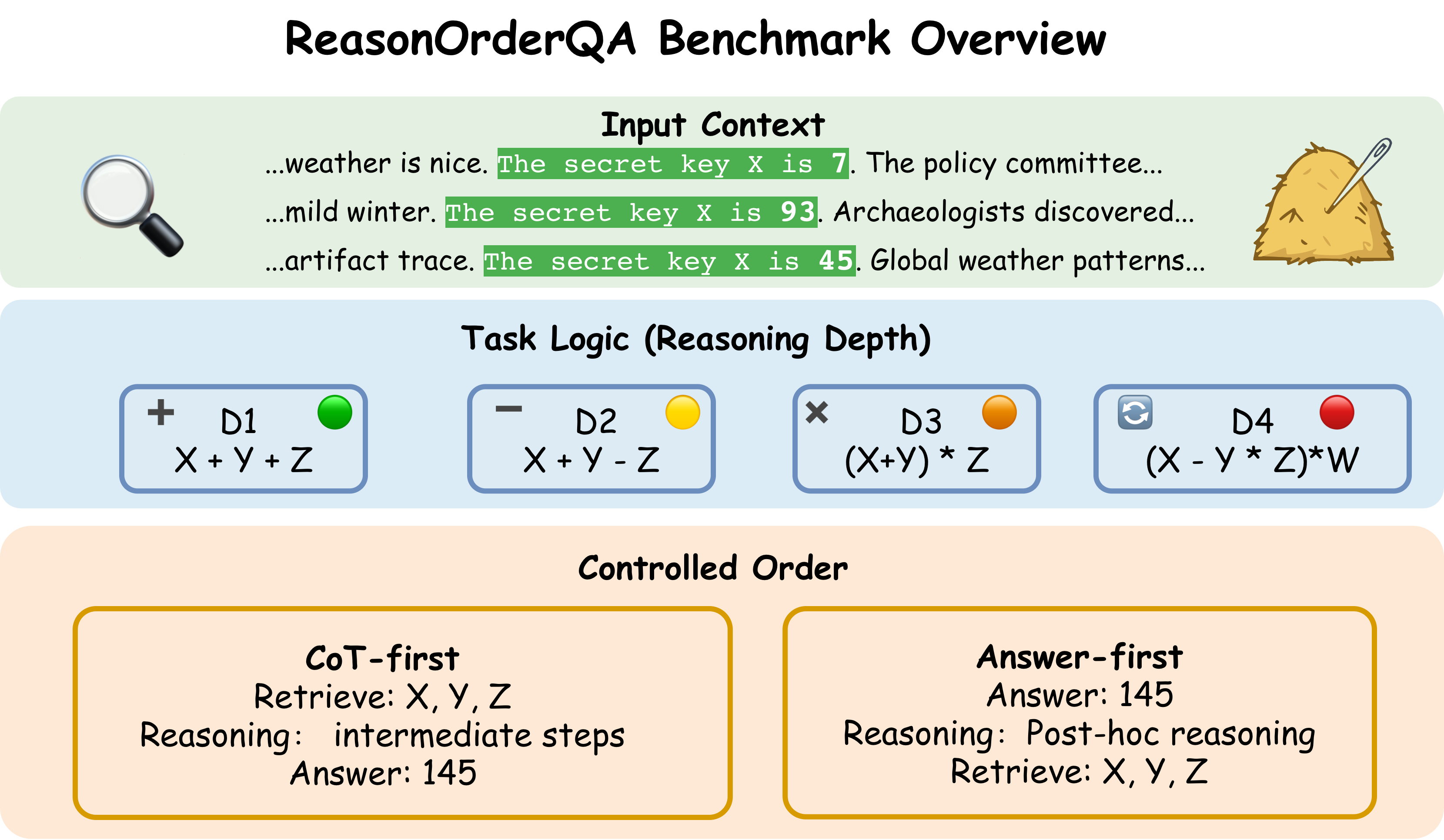}
    \caption{\textbf{ReasonOrderQA benchmark design.} 
    Each problem contains (1) a noisy context with hidden variable definitions, 
    (2) an arithmetic formula of varying complexity (D1-D4), and 
    (3) evaluation under two generation orders (Standard: reasoning→answer, Reverse: answer→reasoning). 
    This design decouples retrieval difficulty from reasoning complexity.}
    \label{fig:orderqa_overview}
\end{figure*}
\paragraph{Masked Diffusion Language Models.} Masked diffusion LMs like LLaDA~\citep{nie2025llada} define a reverse process that recovers clean data from a fully masked sequence. A bidirectional Transformer predicts all masked tokens simultaneously, allowing the model to access global context at any step and theoretically decoupling generation order from textual position.
% \paragraph{Low-Confidence Remasking.}
Starting from a fully masked sequence at step $t=T$, the model iteratively predicts $\hat{\mathbf{x}}_0$ from $\mathbf{x}_t$. At each diffusion step, high-confidence tokens are retained while low-confidence tokens are re-masked for refinement in subsequent steps (i.e., tokens with low confidence are ``remasked''). This prioritizes high-certainty predictions globally, regardless of their position in the sequence. We use this as our default strategy throughout the paper; alternative strategies are compared in \S\ref{subsec:sampling_strategies}.

% \yue{I think this should be in the section after the benchmark, this seems to be experimental setup section; if this is important for defining the benchmark, then move this into the assumption of section 3}
\paragraph{Task Formulation and Metrics.}
To precisely analyze the generation dynamics, we formalize the task settings and define the terminology used throughout our experiments.

We consider two output orders for a question $x_Q$ with rationale $x_R$ and answer $x_A$.
In the CoT-First setting, the model produces $\mathbf{x} = [x_Q, x_R, x_A]$, which matches the standard left-to-right reasoning narrative and the autoregressive training objective.
In the Answer-First setting, the model is constrained to output $\mathbf{x} = [x_Q, x_A, x_R]$.
The prompt templates enforcing these formats are provided in Appendix~\ref{app:prompts}.

\paragraph{Generation Settings.}
We use generation length $L = 256$ and diffusion steps $T = 256$. Because the model estimates the entire sequence at every step, we can track internal predictions separately from the completion text (tokens actually committed). Full settings are in Appendix~\ref{app:exp_settings} (Table~\ref{tab:appendix_settings}).

\begin{figure*}[t] 
    \centering 
  \includegraphics[width=0.95\textwidth]{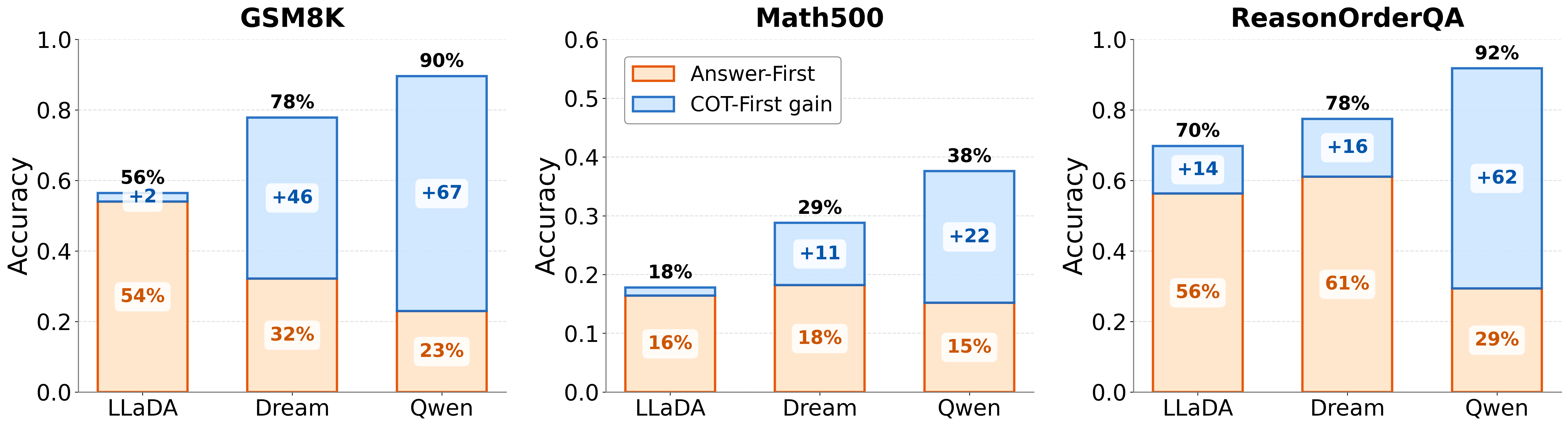}
  \caption{\textbf{Order robustness across diffusion and autoregressive models.} We compare LLaDA (diffusion, trained from scratch), Dream (diffusion, distilled from AR), and Qwen (autoregressive) on GSM8K, Math500, and ReasonOrderQA. Orange bars show Answer-First accuracy; blue portions show additional gain from CoT-First. LLaDA exhibits strong order robustness with minimal CoT-First gain (+2\% on GSM8K). Dream, despite being a diffusion model, shows weaker robustness (+46\% gain), suggesting that distillation from AR models may preserve order-sensitive behavior. Qwen shows the largest gap (+67\%), consistent with AR's structural dependence on generation order.}
    \label{fig:motivation_bars} 
\end{figure*}
\section{ReasonOrderQA Benchmark}
\label{sec:reasonorderqa}

To systematically probe the relationship between generation order and reasoning dynamics, we introduce \textbf{ReasonOrderQA}, a controlled benchmark of 1,000 problems across four difficulty levels: D1 (24\%), D2 (39\%), D3 (26\%), and D4 (11\%). See Appendix~\ref{app:dataset_construction} for construction details.

\paragraph{Limitations of Existing Benchmarks.}
Math500 and GSM8K are suboptimal for studying order robustness: (1) intermediate steps are opaque, flawed reasoning may still yield correct answers; (2) failure modes are conflated and errors could stem from order sensitivity or insufficient capacity; (3) difficulty gradients are weak, e.g. Math500 levels do not produce well-separated behavioral patterns.

\paragraph{Design Goals.}
ReasonOrderQA addresses these limitations with three design goals: (1) Separate retrieval from reasoning: we decouple the retrieval stage (finding hidden variables) from the reasoning stage (computing the formula), enabling independent evaluation—Retrieval F1 measures the fraction of secret keys that appear in the model's output, while Reasoning Accuracy measures whether the final answer is correct; (2) Clear difficulty gradients: four difficulty levels (D1-D4) with explicitly increasing arithmetic complexity, ensuring observable behavioral differences; (3) Verify early convergence for simple tasks: by including trivially simple problems (D1), we can test whether high-confidence tokens indeed converge earlier, validating the confidence-to-order mechanism.

\paragraph{Benchmark Construction.}
As illustrated in Figure~\ref{fig:orderqa_overview}, we embed hidden variable definitions (e.g., ``The secret key X is 7'') within a noisy context filled with distractors. The model must first retrieve these keys and then apply them to an arithmetic formula. We define four difficulty levels based on arithmetic complexity:
\begin{itemize}
    \item D1: $X + Y + Z$ (simple addition)
    \item D2: $X + Y - Z$ (mixed operations)
    \item D3: $(X + Y) \times Z$ (parentheses)
    \item D4: $(X - Y \times Z) \times W$ (nested operations)
\end{itemize}
Each level uses a fixed formula template while numeric values and distractor contexts vary across instances. This setup allows us to track, at each diffusion step, when the model retrieves variables (Retrieval F1) versus when it resolves the final answer (Reasoning Accuracy).

\section{Robustness to Controlled Order}
\label{sec:phenomenon}
This section establishes an empirical baseline for order robustness by comparing autoregressive and diffusion-based models under different output ordering conditions.

\paragraph{Experimental Protocol.} 
We compare three models representing different training paradigms: LLaDA-8B-Instruct, a masked diffusion language model trained from scratch~\citep{nie2025llada}; Dream-7B, a diffusion model distilled from autoregressive weights~\citep{ye2025dream}; and Qwen2.5-7B-Instruct\citep{qwen2.5}, a standard autoregressive model. This comparison allows us to examine whether order robustness stems from the diffusion architecture itself or from training methodology. For diffusion models, we employ the default low-confidence remasking strategy (as described in \S\ref{sec:background}), which prioritizes tokens based on confidence rather than textual position. We evaluate all models on GSM8K, Math500, and ReasonOrderQA under both the standard CoT-First order and the adversarial Answer-First constraint.

As illustrated in Figure~\ref{fig:motivation_bars}, LLaDA exhibits strong order robustness across all three datasets, with accuracy gaps between CoT-First and Answer-First consistently under 15\%. In contrast, Qwen shows substantial order sensitivity, with gaps reaching 22-67\%.

The Dream experiment reveals a key insight. Despite being a diffusion model, Dream shows intermediate behavior with gaps of +46\% (GSM8K), +11\% (Math500), and +16\% (ReasonOrderQA). Dream is distilled from an autoregressive model, and we account this to its intermediate performance. This suggests that order robustness is not an automatic property of diffusion architectures, but also depends on training methodology. Models distilled from AR may inherit order-sensitive representations, partially preserving the coupling between generation order and reasoning order.

\begin{figure}[t]
    \centering
    \includegraphics[width=\columnwidth]{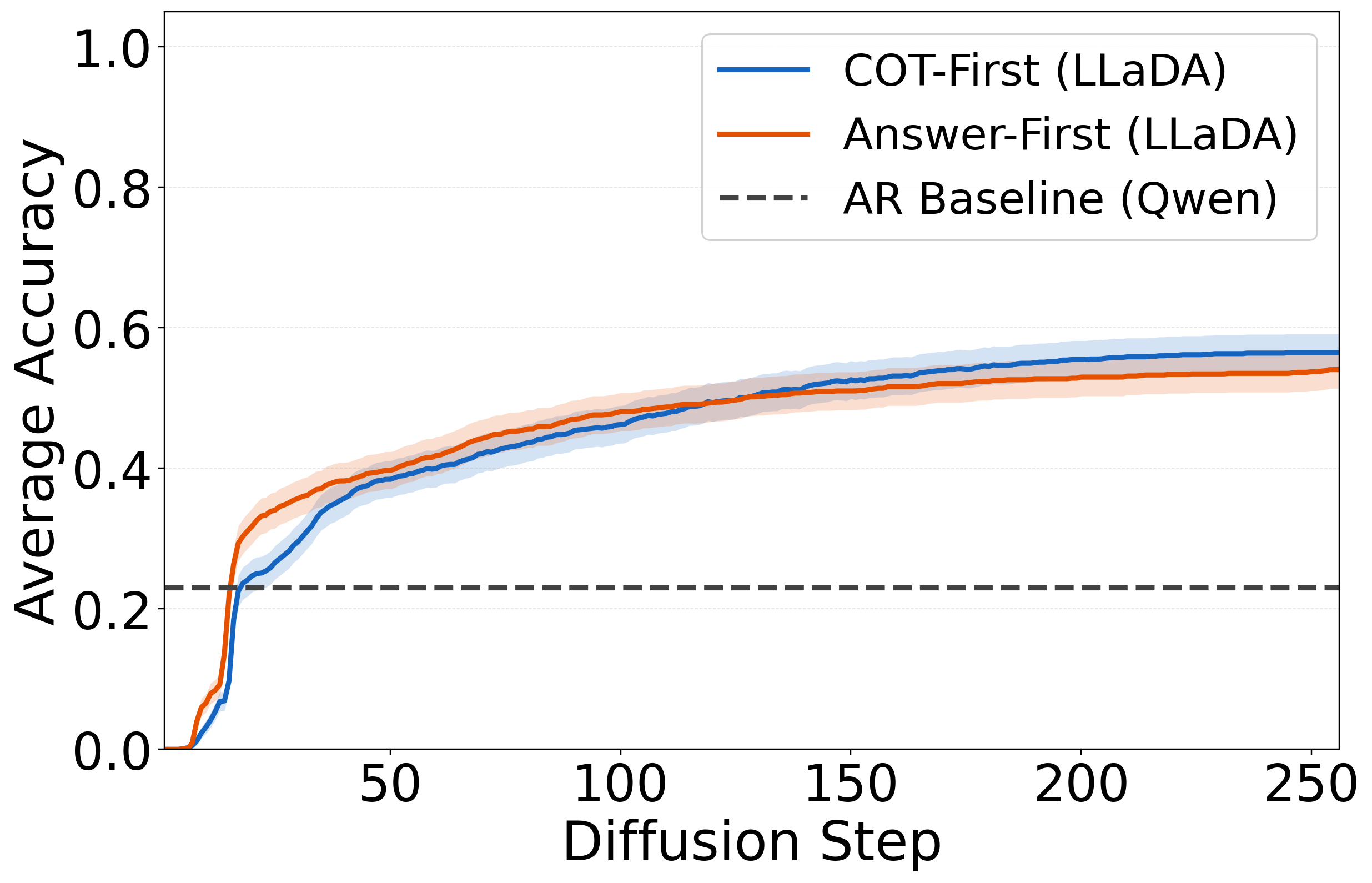} 
    \caption{
        \textbf{Convergence trajectories on GSM8K.}
        Blue and orange lines show LLaDA Diffusion under CoT-First and Answer-First prompting, respectively; gray dashed line shows Qwen2.5-7B final accuracy under Answer-First. 
        Diffusion trajectories overlap almost perfectly regardless of output order, gradually improving to $\sim$55\% accuracy. The Qwen baseline achieves only $\sim$23\% under Answer-First, confirming the AR model's order sensitivity.
    }
    \label{fig:convergence_comparison}
\end{figure}

Figure~\ref{fig:convergence_comparison} further validates this through inference-time dynamics: LLaDA's trajectories overlap regardless of output order, while Qwen degrades substantially. We observe consistent patterns on Math500 (Appendix~\ref{app:trajectory_math500}).

\section{Source of Order Robustness}
\label{sec:mechanism}

Having established the phenomenon, we now analyze the underlying patterns. Prior work has used token-level confidence in autoregressive LMs for uncertainty quantification~\citep{fadeeva2024factchecking, malinin2021uncertainty} and adaptive computation such as early-exiting~\citep{schuster2021consistent, bae2023fast}. However, in AR models, confidence remains a \textit{diagnostic signal}: it does not alter the fixed left-to-right generation order. In diffusion models, this changes fundamentally i.e. confidence directly controls which tokens get unmasked next, turning it into a \textit{generation mechanism}. 

We investigate whether token confidence correlates with task complexity. If harder tokens exhibit systematically lower confidence, the sampling algorithm would naturally defer them, creating an implicit complexity-based ordering regardless of textual position. Our analysis documents this empirical pattern across controlled difficulty levels.

\paragraph{Methodology.} ReasonOrderQA provides an ideal testbed because its difficulty levels (D1-D4) directly manipulate answer token complexity: D1 requires simple addition, while D4 requires nested operations with operator precedence. We track two quantities: (1) token confidence: the probability assigned to the predicted token at each diffusion step, and (2) exposure timing: the step at which all answer tokens are unmasked. The structured output template (Answer: / Reasoning: / Retrieval:) provides natural delimiters for tracking each segment.

\subsection{Task Complexity Reflects in Confidence}

\begin{figure}[t]
    \centering
    \includegraphics[width=0.48\textwidth]{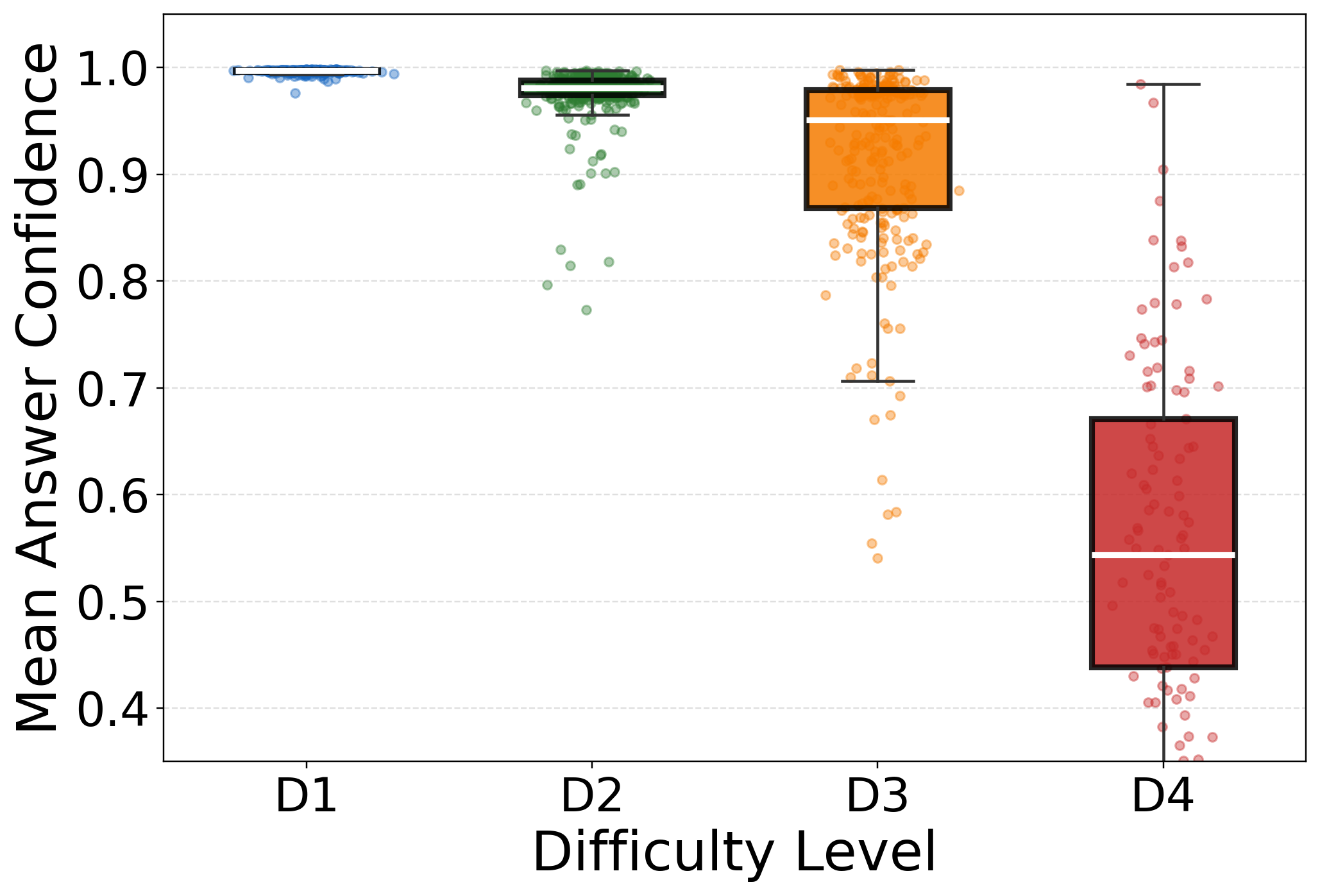}
    \caption{
    \textbf{Answer token confidence decreases with problem difficulty.}
    For each problem, we compute the mean confidence across all diffusion steps for answer tokens (located via the ``Answer:'' delimiter). Each box shows the distribution across problems within that difficulty level. D1 exhibits the highest mean confidence ($\sim$0.99), while D4 shows the lowest ($\sim$0.55). Variance also increases with difficulty. See Figure~\ref{fig:answer_confidence_boxplot} for Math500 results.
    }
    \label{fig:reasonorder_confidence_boxplot}
\end{figure}

Figure~\ref{fig:reasonorder_confidence_boxplot} shows that answer token confidence decreases monotonically with task difficulty. For simple addition (D1), the model achieves near-perfect confidence ($\sim$0.99), indicating that the answer can be resolved almost immediately. As arithmetic complexity increases through D2-D4, confidence drops progressively, reaching only $\sim$0.55 for the most complex nested operations (D4). See Appendix~\ref{app:confidence_trajectories} for step-by-step confidence trajectory visualizations.

Beyond mean confidence, the variance pattern is equally informative. D1-D2 problems cluster tightly around their means, suggesting consistent difficulty across instances. In contrast, D3-D4 exhibit substantial spread, reflecting that some problems within these levels are tractable while others remain challenging. This variance increase indicates that the model's confidence is sensitive not only to the formula structure but also to the specific numeric values involved.

This relationship between task complexity and confidence has a direct implication for generation order: since the sampling algorithm unmasks high-confidence tokens first, simpler answer tokens will be exposed earlier than complex ones, regardless of their textual position. We verify this prediction in the next subsection.

\subsection{Harder Tokens Unmask Later}

\begin{figure}[t]
    \centering
    \includegraphics[width=\columnwidth]{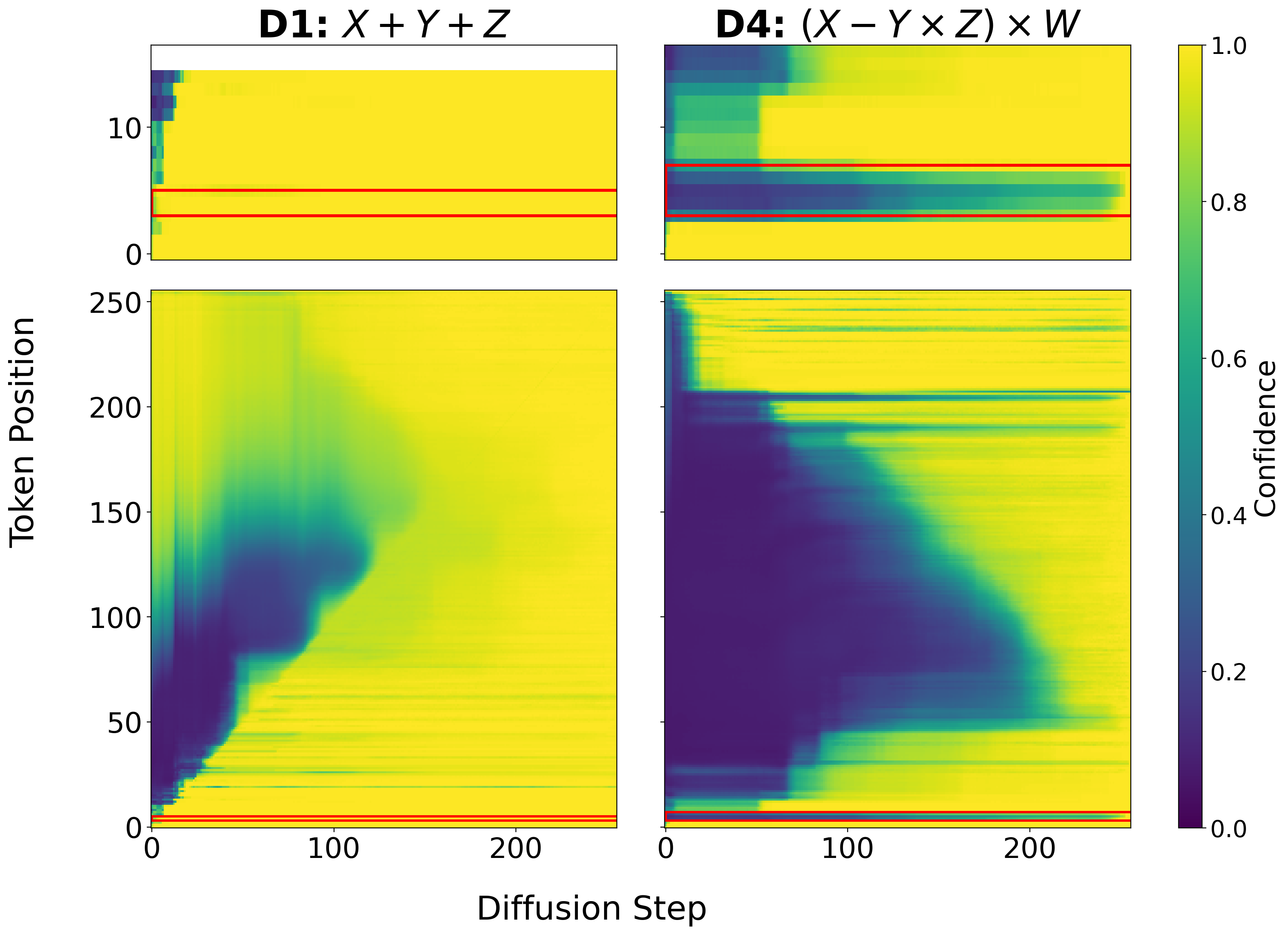}
    \caption{
    \textbf{Confidence dynamics: D1 vs D4.}
    Each heatmap shows token confidence (color) across diffusion steps (x-axis) and token positions (y-axis). Top row zooms into answer tokens (red box); bottom row shows all tokens.
    \textbf{Left (D1):} Answer tokens achieve high confidence (yellow) within $\sim$10 steps.
    \textbf{Right (D4):} Answer tokens remain low confidence (blue/green) until $\sim$200 steps, while many other tokens also show delayed convergence.
    }
    \label{fig:confidence_heatmap}
\end{figure}

Figure~\ref{fig:confidence_heatmap} reveals how confidence dynamics differ across difficulty levels. First, harder tasks produce more low-confidence tokens overall: in D1, most tokens converge quickly (bottom panel shows predominantly yellow), while D4 reveals many more low-confidence tokens (blue/green regions) across the entire generation. Second, since the sampling algorithm prioritizes high-confidence tokens for unmasking, these patterns directly determine exposure timing: in D1, the answer token region (red box) turns yellow early, so these tokens are unmasked quickly; in D4, the answer region remains blue/green for most of the trajectory, so the algorithm keeps selecting other high-confidence tokens first, delaying answer exposure until $\sim$200 steps.

To further understand how the sampling algorithm drives this behavior, we compare \texttt{low\_confidence} (which prioritizes high-confidence tokens) against \texttt{left\_to\_right} (which forces LLaDA to unmask tokens strictly left-to-right while keeping the same model architecture; see Appendix~\ref{app:decoding_formulation}). This comparison isolates the effect of sampling strategy. Table~\ref{tab:reasonorder_answer_exposure} quantifies this effect. The relationship is notably non-linear where, D1-D2 show relatively fast exposure (10-13 steps), but D3 jumps to 63.1 and D4 to 203.1 steps. The total $\Delta$ is +193.0 steps for \texttt{low\_confidence} versus +3.4 for \texttt{left\_to\_right} (57$\times$ difference), confirming that low-confidence tokens are delayed only when the sampling strategy allows confidence-based reordering.

\begin{table}[t]
    \centering
    \small
    \caption{
    \textbf{Answer token exposure timing across ReasonOrderQA complexity levels.}
    \texttt{low\_confidence} delays answer exposure as complexity increases ($\Delta$=+193.0 steps from D1 to D4), while \texttt{left\_to\_right} exposes answers at fixed early steps regardless of complexity ($\Delta$=+3.4 steps). See Appendix~\ref{app:math_exposure} for Math500 results.
    }
    \label{tab:reasonorder_answer_exposure}
    \resizebox{0.95\linewidth}{!}{
    \begin{tabular}{llcc}
    \toprule
    \textbf{Level} & \textbf{Formula} & \textbf{\texttt{left\_to\_right}} & \textbf{\texttt{low\_conf.}} \\
    \midrule
    D1 & $X + Y + Z$ & 10.0 & 10.1 \\
    D2 & $X + Y - Z$ & 10.1 & 12.8 \\
    D3 & $(X + Y) \times Z$ & 11.9 & \textbf{63.1} \\
    D4 & $(X - Y \times Z) \times W$ & 13.4 & \textbf{203.1} \\
    \bottomrule
    \end{tabular}}
    
\end{table}

\begin{figure*}[t]
    \centering
    \includegraphics[width=\textwidth]{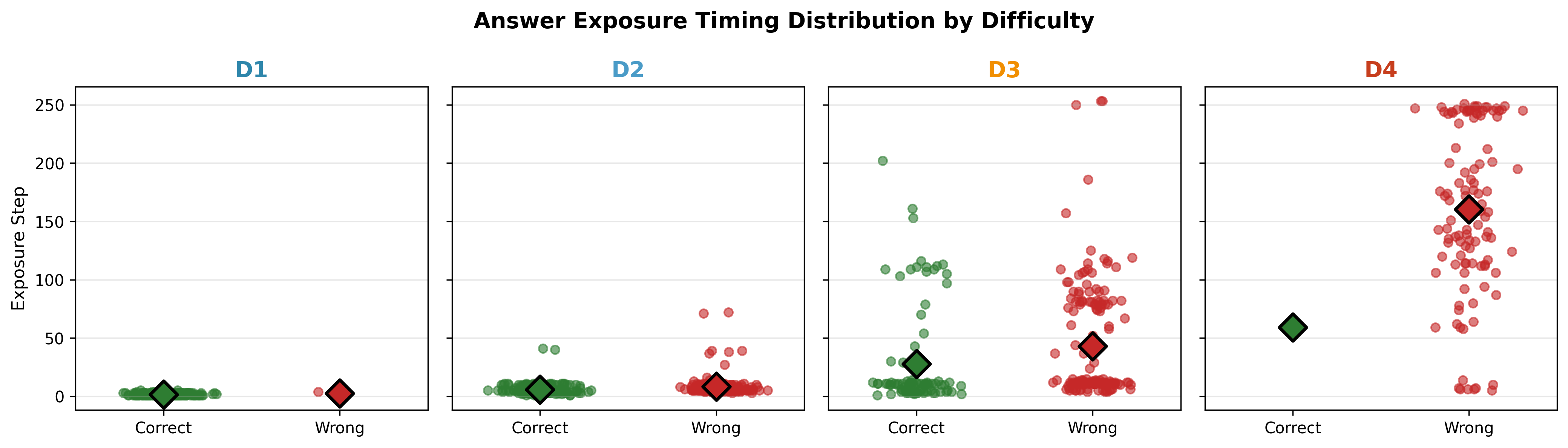}
    \caption{
    \textbf{Answer exposure timing distribution by difficulty level.}
    The \textit{exposure step} is the first diffusion step at which all answer tokens are unmasked (i.e., no mask tokens remain in the answer region).
    Each point represents one problem; green indicates correct answers, red indicates wrong answers. Diamond markers show the mean exposure step for each group. As difficulty increases (D1$\to$D4), correct answers maintain consistent exposure timing patterns, while wrong answers exhibit increasingly scattered and extreme exposure steps. In D4, the single correct case (n=1) has notably lower exposure step than the wrong cases, suggesting that appropriate delay correlates with success.
    }
    \label{fig:exposure_scatter}
    \vspace{-1em}
\end{figure*}

\subsection{Complexity-Based Ordering Enables Correctness}

The previous subsections document a consistent empirical pattern: task complexity correlates with token confidence, and low-confidence tokens are deferred by the sampling algorithm. 
This complexity-correlated ordering appears to enable order robustness. 
By deferring complex answer tokens, the model gains sufficient refinement time to resolve them correctly, regardless of their textual position.

Figure~\ref{fig:exposure_scatter} provides evidence for this pattern. Correct answers follow expected exposure timing, while wrong answers deviate. In D1-D2, both correct and wrong answers cluster at low exposure steps, as the tasks are simple enough to resolve quickly. However, in D3, correct answers maintain a consistent exposure distribution (mean $\sim$30 steps), while wrong answers show higher variance and more extreme values. In D4, where only one answer is correct, that single success has a notably lower exposure step than the failures.

This reveals why diffusion models achieve order robustness: confidence-based remasking implicitly prioritizes reasoning before answer commitment by delaying complex tokens until sufficient refinement occurs. This pattern holds when complexity differences are clear (D3), it breaks down when tokens are exposed too early (premature commitment) or the model never reaches sufficient confidence (D4). This explains both the robustness observed in \S\ref{sec:phenomenon} and sets up our investigation of breakdown conditions in the next section. We provide an information-theoretic formalization connecting this mechanism to recent analyses of parallel decoding in Appendix~\ref{app:info_theory}.

\section{When Order Robustness Breaks Down}
\label{sec:benchmarking}

The evidence in \S\ref{sec:mechanism} suggests that token ordering correlates with relative complexity, but when does this pattern break down? We identify two key factors: (1) insufficient complexity differences between tokens, and (2) large generation lengths. Using ReasonOrderQA under the Answer-First setting, we test these factors by tracking the temporal relationship between retrieval, reasoning, and answer tokens.

\begin{figure}[t]
    \centering
    \includegraphics[width=\linewidth]{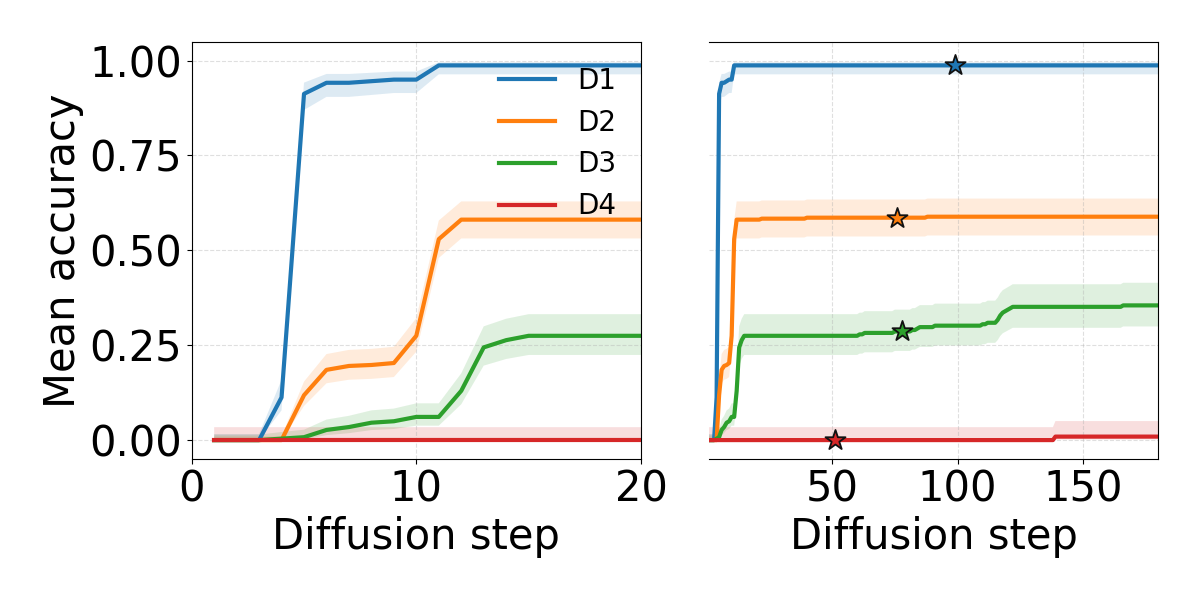}
    \caption{\textbf{Mean Accuracy under Answer-First across difficulty levels.} Left: early-stage dynamics (steps 0-20). Right: full trajectory (steps 0-200). Stars mark the step at which retrieval accuracy reaches 0.95, indicating when the model has internally identified the relevant variables. D1 converges quickly; D2 and D3 show gradual improvement; D4 remains near zero ($\sim$1\%, 1/109 correct) as the task exceeds model capacity.}
    \label{fig:figure6_completion_text_relax}
    \vspace{-1em}
\end{figure}

\subsection{Breakdown Condition 1: Insufficient Complexity Differences}
\label{subsec:trajectory_analysis}

Figure~\ref{fig:figure6_completion_text_relax} shows how accuracy evolves during Answer-First generation. We observe that order robustness breaks down when tokens have similar complexity levels. 

The contrast between $D_2$ and $D_3$ is particularly revealing. In $D_2$ (mixed addition/subtraction), accuracy plateaus early at $\sim$55\% and shows minimal improvement after step 20. The answer tokens ($X + Y - Z$) and retrieval tokens (finding X, Y, Z values) have similar complexity, so the model cannot reliably distinguish which to decode first. This leads to early commitment that results in errors that cannot be corrected in later steps. In contrast, $D_3$ (parenthesized multiplication) demonstrates successful robustness: accuracy continues rising from $\sim$25\% at step 20 to $\sim$35\% at convergence. The stars in Figure~\ref{fig:figure6_completion_text_relax} mark when retrieval converges (0.95 F1); crucially, $D_3$ shows substantial \textit{post-retrieval} accuracy growth after this point, while $D_2$ has already plateaued. This indicates that $D_3$'s retrieval tokens require only pattern matching while its answer tokens require multi-step computation, creating a complexity gap that allows the model to correctly prioritize retrieval before committing to the answer.

Notably, $D_4$ shows a different pattern: it remains near zero ($\sim$1\%) throughout, indicating that $D_4$ exceeds the model's reasoning capacity, implying that the task itself is too difficult regardless of ordering. $D_1$ converges almost immediately to near-perfect accuracy ($>$95\%), confirming that simple tasks pose no ordering challenge. This finding reveals that \textit{complexity differences} between tokens within the same problem, not problem-level complexity, determine when robustness breaks down.

\subsection{Breakdown Condition 2: Large Generation Length}
\label{subsec:block_length}

\begin{figure}[t]
    \centering
    \includegraphics[width=\linewidth]{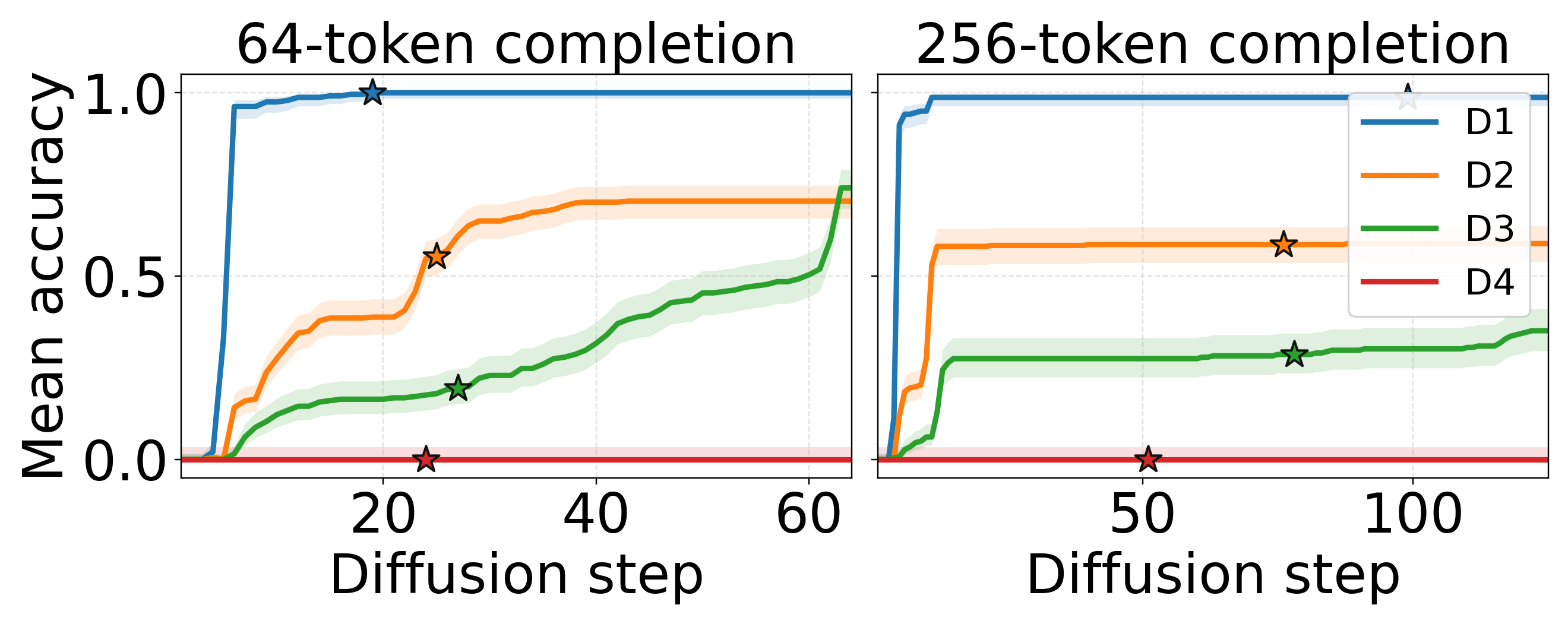}
    \caption{Comparison of reasoning accuracy between 64-token and 256-token generation lengths. Shorter lengths demonstrate superior post-retrieval reasoning growth, while longer lengths exhibit earlier performance saturation.}
    \label{fig:comparison_window_length}
\end{figure}

\begin{table}[t]
    \centering
    \footnotesize
    \caption{
    \textbf{Latent Retrieval Availability.} Steps required to reach 0.95 Retrieval F1 in internal predictions. With 64 tokens, all difficulties achieve high F1 within 4-11 steps. With 256 tokens, retrieval is significantly delayed for D1-D3, while D4 remains fast due to its distinctive formula structure.
    }
    \label{tab:retrieval_prediction}
    \begin{tabular}{@{}llccc@{}}
    \toprule
    \textbf{Level} & \textbf{Formula} & \textbf{64} & \textbf{256} & \textbf{$\Delta$} \\
    \midrule
    D1 & $X\!+\!Y\!+\!Z$ & 11 & \textbf{87} & +76 \\
    D2 & $X\!+\!Y\!-\!Z$ & 4 & \textbf{49} & +45 \\
    D3 & $(X\!+\!Y)\!\times\!Z$ & 8 & \textbf{63} & +55 \\
    D4 & $(X\!-\!Y\!\times\!Z)\!\times\!W$ & 9 & 20 & +11 \\
    \bottomrule
    \end{tabular}
    \vspace{-1em}
\end{table}

As illustrated in Figure~\ref{fig:comparison_window_length}, larger generation length ($L_{\text{gen}} = 256$ vs $L_{\text{gen}} = 64$) significantly degrades reasoning accuracy. With 64 tokens, the model maintains post-retrieval growth (reasoning accuracy continues rising after retrieval stabilizes), but with 256 tokens, this growth is suppressed. This suggests that the number of tokens the model must simultaneously predict may limit its ability to judge complexity differences. This limitation is particularly evident in $D_2$, where tokens have similar complexity levels.

To further verify this, we examine when retrieval, the simplest subtask requiring only pattern matching, converges in the model's internal predictions. Since diffusion models predict all tokens simultaneously, we can inspect their latent beliefs at each step. Table~\ref{tab:retrieval_prediction} shows the steps required to achieve 0.95 Retrieval F1. With 64 tokens, retrieval is immediate (all difficulties reach the threshold within 4-11 steps). With 256 tokens, retrieval is significantly delayed: $D_1$ requires 87 steps, $D_2$ requires 49 steps, and $D_3$ requires 63 steps, while $D_4$ remains fast (20 steps) due to its distinctive formula structure. This confirms that larger generation lengths make it harder for the model to resolve even simple token predictions, delaying recognition and ultimately suppressing post-retrieval reasoning growth.

\paragraph{The Generation Length Trade-off.} LLaDA requires specifying the maximum generation length upfront, yet our findings show that larger lengths degrade order robustness. This creates a dilemma: complex problems requiring longer CoT reasoning need larger generation lengths, but this very increase undermines the model's ability to distinguish token complexity. Resolving this, either through adaptive or block-wise generation is a promising direction for future work.

\subsection{Impact of Sampling Strategies}
\label{subsec:sampling_strategies}

Our analysis above uses the low-confidence remasking strategy, which prioritizes high-confidence tokens for unmasking. How do different sampling strategies affect order robustness? We evaluate multiple strategies (detailed in Appendix~\ref{app:decoding_formulation}) under both CoT-First and Answer-First conditions on ReasonOrderQA.

\definecolor{dropveryhard}{HTML}{D73027}
\definecolor{drophard}{HTML}{FC8D59}
\definecolor{dropmedium}{HTML}{FEE08B}
\definecolor{dropmild}{HTML}{D9EF8B}
\definecolor{droplow}{HTML}{91CF60}
\definecolor{dropverylow}{HTML}{1A9850}
\definecolor{arbaseline}{HTML}{E0E0E0}

\begin{table}[t]
    \centering
    \small
    \caption{
    \textbf{Accuracy (\%) across sampling strategies and output orders on ReasonOrderQA.}
    Sorted by Answer-First accuracy. We compare Qwen2.5-7B (AR) as a baseline against various diffusion sampling strategies. The AR baseline shows a substantial 68\% relative drop under Answer-First prompting. Among diffusion strategies, \texttt{topk\_margin} achieves the highest Answer-First accuracy (60.0\%), while \texttt{entropy} achieves the best order robustness (-9.8\% relative). Color intensity indicates relative performance drop.
    }
    \label{tab:sampling_strategies}
    \begin{tabular}{lccc}
    \toprule
    \textbf{Strategy} & \textbf{CoT-First} & \textbf{Answer-First} & \textbf{$\Delta$ (rel)} \\
    \midrule
    \texttt{topk\_margin} & 74.7 & \textbf{60.0} & \cellcolor{dropmedium}-19.7\% \\
    \texttt{low\_confidence} & 69.0 & 57.3 & \cellcolor{dropmild}-17.0\% \\
    \texttt{entropy} & 60.4 & 54.5 & \cellcolor{droplow}-9.8\% \\
    \texttt{random} & 60.1 & 51.1 & \cellcolor{dropmild}-15.0\% \\
    \texttt{left\_to\_right} & 64.0 & 47.2 & \cellcolor{drophard}-26.3\% \\
    \midrule
    \rowcolor{arbaseline} \textit{Qwen2.5-7B (AR)} & \textit{91.8} & \textit{29.4} & \cellcolor{dropveryhard}\textit{-68.0\%} \\
    \bottomrule
    \end{tabular}
\end{table}

Table~\ref{tab:sampling_strategies} reveals two key findings. First, all diffusion strategies substantially outperform the AR baseline in order robustness: Qwen2.5-7B achieves the highest CoT-First accuracy (91.8\%) but shows a substantial 68\% relative drop under Answer-First prompting, while all diffusion strategies show much smaller drops (9.8\%-26.3\%). Second, Answer-First accuracy reveals a clear hierarchy based on the information used for unmasking: \texttt{left\_to\_right} achieves only 47.2\%, \texttt{random} reaches 51.1\%, while uncertainty-based strategies progressively improve \texttt{entropy} (54.5\%), \texttt{low\_confidence} (57.3\%), and \texttt{topk\_margin} (60.0\%). 

This hierarchy reveals a key insight: uncertainty-based strategies all derive their signal from the same source i.e. the token probability distribution whether summarized as top-1 confidence, full-distribution entropy, or top-1 vs.\ top-2 margin. Access to this distributional information is what enables order robustness. The variation among uncertainty-based methods reflects a secondary accuracy-robustness trade-off: aggressive unmasking (\texttt{topk\_margin}) achieves higher accuracy but risks premature commitment, while conservative strategies (\texttt{entropy}) preserve flexibility at the cost of absolute performance.

\section{Related Work}
\label{sec:related_work}

We situate our work at the intersection of generation paradigms, diffusion models, and reasoning mechanisms.

\paragraph{Sequential vs. Parallel Generation.}
Autoregressive (AR) models generate tokens strictly left-to-right, creating a structural bias that limits global planning~\citep{vaswani2017attention, brown2020language}. This constraint causes failures in non-monotonic tasks and bidirectional reasoning, such as the reversal curse~\citep{berglund2023reversal, welleck2019nonmonotonic}. Non-autoregressive (NAR) approaches attempt to break this linearity through parallel decoding~\citep{gu2017nonautoregressive}, iterative refinement~\citep{lee2018deterministic, ghazvininejad2019maskpredict}, or insertion-based strategies~\citep{gu2019indigo, chan2020imputer}. These methods, however, typically operate within fixed windows and do not fundamentally decouple reasoning order from generation order.

\paragraph{Diffusion Models for Text.}
Diffusion models introduce a parallel denoising paradigm~\citep{austin2021structured, li2022diffusionlm}. Recent work has scaled this approach to large language models (LLMs) via discrete or masked diffusion~\citep{lou2023discrete, shi2024simplified, nie2024scaling, sahoo2024simple}, with LLaDA~\citep{nie2025llada} achieving state-of-the-art performance. While some studies explore adaptation from AR weights~\citep{gong2025scaling, gulrajani2023likelihood} or block-wise generation~\citep{arriola2025blockdiffusion}, most focus on final output quality~\citep{yi2024diffusiontextsurvey}. Our work differs by probing the internal \textit{dynamics} of generation to reveal how order invariance emerges.

\paragraph{Order Robustness and Reasoning.}
Prior work examines the reversal curse~\citep{berglund2023reversal}, non-monotonic generation~\citep{welleck2019nonmonotonic}, and chain-of-thought robustness under noise~\citep{zhou2024noisyrationales, wang2025cotbounds}. Recent work by \citet{chen2025beyond} formalizes the Parallel-Sequential Contradiction (PSC), showing that DLLMs may revert to autoregressive-like behavior as task complexity increases. Our work offers a complementary perspective: confidence-based remasking can \textit{leverage} token dependencies to achieve order-invariant reasoning.

\section{Conclusion}
\label{sec:conclusion}

We show that masked diffusion language models exhibit order robustness: unlike AR models that degrade substantially under Answer-First prompting (up to 67\% relative drop), diffusion models trained from scratch remain stable ($\leq$4\% relative drop). 
The underlying mechanism is complexity-based ordering where confidence-based sampling defers complex tokens until sufficient refinement occurs. 
Through ReasonOrderQA, we identify two breakdown conditions: insufficient complexity differences between tokens, and large generation lengths. 
In the ongoing debate between autoregressive and diffusion language models, we identify a scenario where diffusion-style generation is preferable: when the required output order conflicts with the model's natural reasoning order. This provides a criterion for model selection and a design signal for future architectures aiming to balance instruction-following with accurate reasoning.

\section*{Impact Statement}
This paper presents work whose goal is to advance the field of Machine Learning. There are many potential societal consequences of our work, none which we feel must be specifically highlighted here.

\bibliographystyle{icml2026}
\bibliography{custom}

\appendix
\crefalias{section}{appendix}
\crefalias{subsection}{appendix}
\crefalias{subsubsection}{appendix}
\label{sec:appendix}
\onecolumn
\newpage

\section{Experimental Setup}
\label{app:exp_settings}

Table~\ref{tab:appendix_settings} outlines the hyperparameter configuration used for both the \textsc{CoT-first} and \textsc{Answer-first} experimental conditions. We evaluate three models: \textsc{LLaDA}-8B-Instruct (diffusion, trained from scratch), \textsc{Dream}-7B (diffusion, distilled from AR), and \textsc{Qwen2.5-7B-Instruct} (autoregressive). All models share the same generation settings except where noted. For autoregressive decoding (Qwen), we set max tokens to 512 and temperature to 0. For \textsc{Dream}, we use temperature $= 0.1$ instead of 0, as temperature 0 causes the model to collapse into degenerate outputs; all other settings remain identical to \textsc{LLaDA}. To ensure reproducibility, deterministic decoding is employed for all models.

\begin{table}[ht]
    \centering
    \caption{Hyperparameters for the experimental evaluation. We set $L_{\text{gen}} = T = L_{\text{block}} = 256$, following the recommended configuration where the stepwise decoding rate ($L_{\text{gen}}/T = 1$) ensures one token is unmasked per diffusion step, which balances quality and efficiency.}
    \label{tab:appendix_settings}
    \renewcommand{\arraystretch}{1.3}
    \begin{tabular}{ll} 
        \toprule
        \textbf{Parameter} & \textbf{Value} \\
        \midrule
        Model & \textsc{LLaDA}-8B-Instruct \\
        Gen. Length ($L_{\text{gen}}$) & 256 \\
        Block Size ($L_{\text{block}}$) & 256 \\
        Diff. Steps ($T$) & 256 \\
        Stepwise Rate ($L_{\text{gen}}/T$) & 1 \\
        Strategy & Deterministic (temp.\ $= 0$) \\
        Remasking & Low-confidence \\
        Hardware & NVIDIA A100 \\
        \bottomrule
    \end{tabular}
\end{table}

\section{ReasonOrderQA Dataset Construction}
\label{app:dataset_construction}

We describe the generation procedure for ReasonOrderQA, a controlled benchmark designed to probe the relationship between generation order and reasoning dynamics. The dataset contains \textbf{1,000 problems} with an approximate target passage length of 1,000 tokens per problem.

\subsection{Difficulty Sampling}

Problems are sampled according to the distribution: D1:D2:D3:D4 = 0.25:0.40:0.25:0.10. This ensures coverage across all complexity levels while reflecting the natural distribution where moderate-difficulty problems (D2) are most common.

\subsection{Expression Generation}

Each difficulty level uses a fixed arithmetic expression template with randomly sampled integers:

\begin{itemize}[leftmargin=2cm]
    \item \textbf{D1:} $X + Y + Z$ where $X, Y, Z \in [1, 20]$
    \item \textbf{D2:} $X + Y - Z$ where $X, Y, Z \in [1, 50]$
    \item \textbf{D3:} $(X + Y) \times Z$ where $X, Y, Z \in [1, 100]$
    \item \textbf{D4:} $(X - Y \times Z) \times W$ where $X, Y, Z, W \in [1, 100]$
\end{itemize}

The gold answer is computed directly from the expression. D1-D2 use three variables, while D3-D4 introduce parentheses and D4 adds a fourth variable, creating strict complexity gradients.

\subsection{Passage Generation}

Each problem includes a passage of approximately 1000 tokens that embeds the secret key values. The passage is constructed as follows:

\begin{enumerate}[leftmargin=2.5cm]
    \item \textbf{Filler content:} Random sentences are sampled from a pool of distractor sentences and filler fragments. Some sentences are extended with 3-8 additional fragment words to vary length.
    
    \item \textbf{Key insertion:} Sentences containing the secret keys (e.g., ``The secret key X is 42'') are randomly inserted into the passage at different positions.
    
    \item \textbf{Length normalization:} Additional filler content is appended until the passage reaches the target length (1000 tokens).
\end{enumerate}

This design ensures that key retrieval requires scanning the entire passage rather than relying on positional heuristics.

\subsection{Retrieval F1 Computation}

Retrieval F1 measures the fraction of ground-truth secret keys that appear in the model's output. For ReasonOrderQA, each problem has a set of gold keys $\mathcal{K} = \{k_1, k_2, \ldots, k_n\}$ (where $n=3$ for D1-D3 and $n=4$ for D4).

\paragraph{Delimiter Detection.}
We detect the ``Retrieval:'' segment using string matching on the output text. The segment is defined as all text between the ``Retrieval:'' delimiter and either the next delimiter (``Reasoning:'' or ``Answer:'') or the end of the generation. For internal predictions (before tokens are committed), we apply the same procedure to the model's predicted sequence $\hat{\mathbf{x}}_0^{(t)}$ at each diffusion step $t$.

\paragraph{Key Extraction.}
From the detected retrieval segment, we extract all integers using regular expression matching (\texttt{$\backslash$d+}). Let $\mathcal{P}$ denote the set of extracted integers from the model's output.

\paragraph{F1 Calculation.}
We compute precision, recall, and F1 as follows:
\begin{align}
    \text{Precision} &= \frac{|\mathcal{P} \cap \mathcal{K}|}{|\mathcal{P}|}, \quad
    \text{Recall} = \frac{|\mathcal{P} \cap \mathcal{K}|}{|\mathcal{K}|} \\[0.3cm]
    \text{F1} &= \frac{2 \cdot \text{Precision} \cdot \text{Recall}}{\text{Precision} + \text{Recall}}
\end{align}

When $|\mathcal{P}| = 0$ (no integers extracted), we set Precision $= 0$ and F1 $= 0$.

\paragraph{Handling Partial Matches.}
We use exact integer matching: a predicted key matches a gold key if and only if they are numerically identical. Order is not considered, and we treat both $\mathcal{P}$ and $\mathcal{K}$ as sets. Duplicate predictions are counted only once (i.e., predicting the same key twice does not inflate precision).

\subsection{Reasoning Accuracy}

Reasoning Accuracy is a binary metric: 1 if the model's final answer exactly matches the gold answer, 0 otherwise. The answer is extracted from the ``Answer:'' segment using the same delimiter detection procedure. For answers wrapped in \verb|\boxed{}|, we extract the content inside the box.

\subsection{Exposure Timing}

The \textbf{exposure step} for a segment (Answer, Reasoning, or Retrieval) is defined as the first diffusion step $t$ at which all token positions within that segment are unmasked (i.e., no mask tokens remain in the segment). We detect segment boundaries using the delimiter tokens and track the mask status of each position across diffusion steps.

\section{Math500 Analysis}
\label{app:trajectory_math500}
\label{app:math_exposure}

\begin{figure}[ht]
    \centering
    \includegraphics[width=0.5\textwidth]{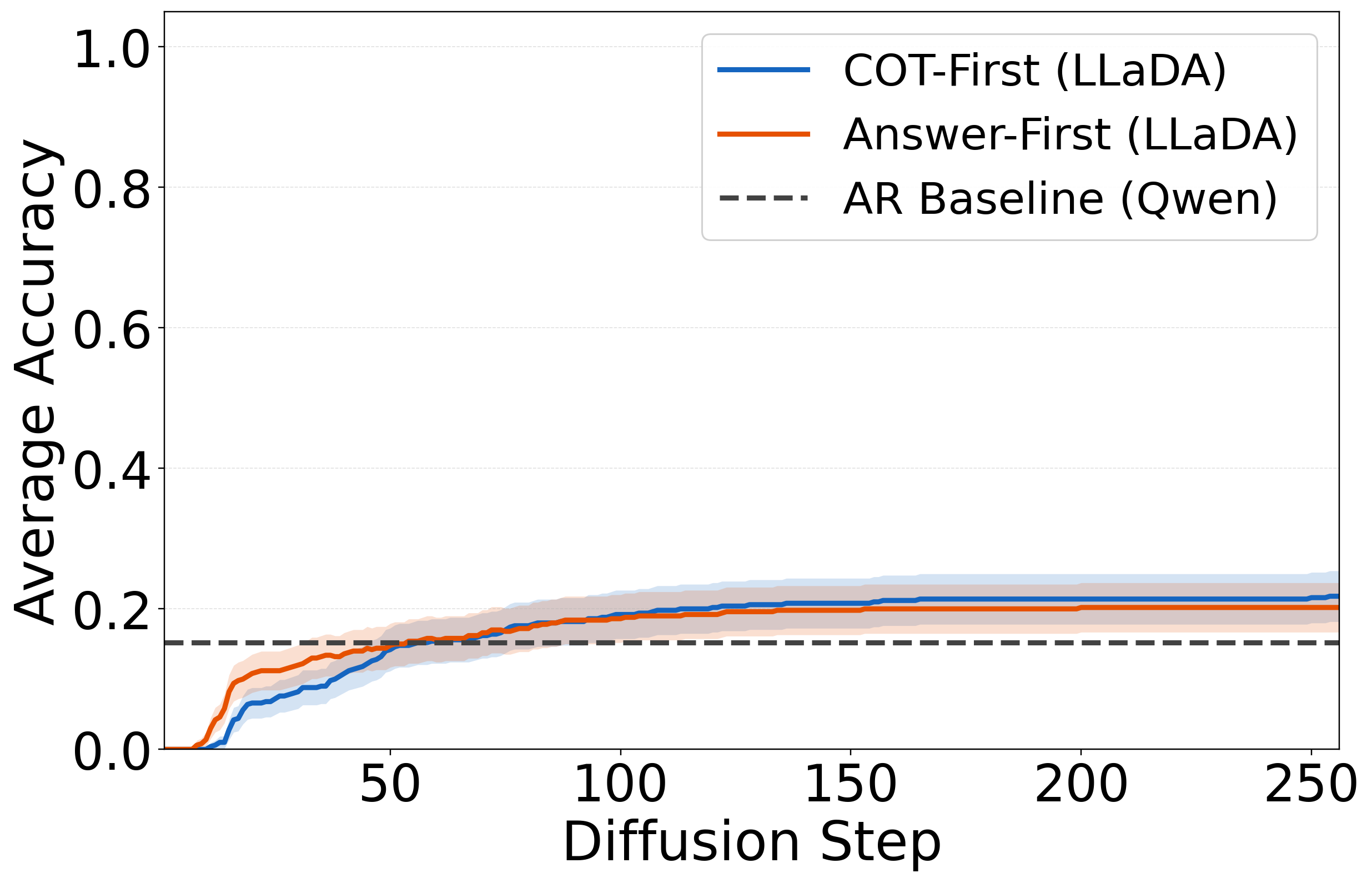} 
    \caption{
        \textbf{Convergence trajectories on Math500.}
        Blue and orange lines show LLaDA Diffusion under CoT-First and Answer-First prompting, respectively; gray dashed line shows Qwen2.5-7B under Answer-First.
        Diffusion trajectories overlap almost perfectly regardless of output order, gradually improving to $\sim$20\% accuracy. Qwen remains flat at $\sim$15\%, consistent with the patterns observed on GSM8K (see Figure~\ref{fig:convergence_comparison}).
    }
    \label{fig:convergence_comparison_math500}
\end{figure}

We observe consistent trajectory patterns on Math500, confirming that the order robustness observed on GSM8K generalizes to other reasoning benchmarks. Figure~\ref{fig:convergence_comparison_math500} compares the evolution of answer prediction accuracy throughout the generation process. The trajectories show the same fundamental pattern: Qwen (gray dashed) remains flat at $\sim$15\% under Answer-First, while LLaDA Diffusion (blue and orange) demonstrates gradual convergence to $\sim$20\% regardless of output order.

\begin{table}[t]
    \centering
    \small
    \caption{
    \textbf{Answer token exposure timing across Math500 difficulty levels.}
    Diffusion (low-confidence remasking) shows moderate adaptation to difficulty ($\Delta$=+22.4 steps from Level 1 to Level 5), while \texttt{left\_to\_right} maintains nearly constant early exposure ($\Delta$=+1.2 steps). Math500 exhibits less dramatic difficulty-based adaptation compared to ReasonOrderQA (Table~\ref{tab:reasonorder_answer_exposure}), likely due to its more uniform problem structure.
    }
    \label{tab:math500_answer_exposure}
    \begin{tabular}{lcc}
    \toprule
    \textbf{Level} & \textbf{\texttt{left\_to\_right}} & \textbf{low\_confidence} \\
    \midrule
    1 & 12.5 & 49.3 \\
    2 & 12.4 & 49.2 \\
    3 & 13.1 & 68.2 \\
    4 & 12.9 & 73.1 \\
    5 & 13.6 & 71.6 \\
    \bottomrule
    \end{tabular}
\end{table}

\begin{figure}[ht]
    \centering
    \includegraphics[width=0.5\textwidth]{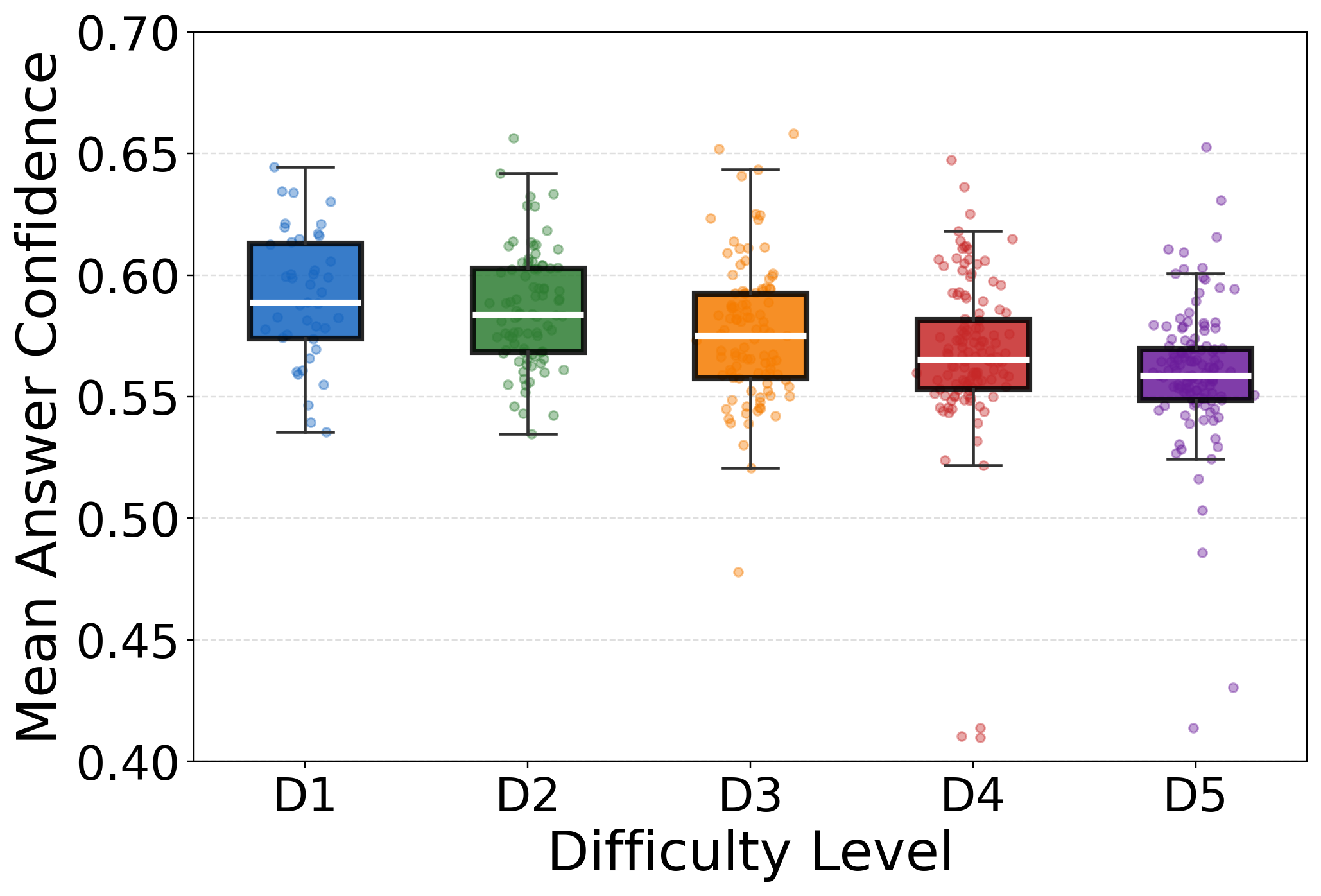}
    \caption{
    \textbf{Answer token confidence decreases with problem difficulty on Math500.}
    For each problem, we compute the mean confidence across all diffusion steps and answer token positions. Each box shows the distribution of per-problem mean confidence within that difficulty level (n = number of problems). As problem complexity increases from Level 1 to Level 5, answer token confidence decreases monotonically.
    }
    \label{fig:answer_confidence_boxplot}
\end{figure}

We examine answer token exposure timing across different problem complexities using Math500, which provides a well-calibrated difficulty spectrum (Level 1 to Level 5). We track the diffusion step at which the Answer segment no longer contains any masked tokens, indicating that the answer has been fully committed to the surface form. Table~\ref{tab:math500_answer_exposure} shows that LLaDA diffusion delays answer token exposure as problem complexity increases: Level 1-2 answers are exposed around step 49, while Level 3-5 answers require approximately 68-73 steps. In contrast, \texttt{left\_to\_right} exposes answer tokens at a fixed early step (~12-13) regardless of problem complexity. 

To further validate this relationship, we directly measure the mean confidence of answer tokens across all diffusion steps. Figure~\ref{fig:answer_confidence_boxplot} shows that answer token confidence decreases monotonically with problem difficulty: Level 1 problems exhibit the highest mean confidence (~0.59), while Level 5 problems show the lowest (~0.56). This confirms that \textbf{task complexity is directly reflected in token-level confidence}, providing the mechanistic basis for why more difficult problems require more diffusion steps before answer tokens can be confidently exposed.

\section{Confidence Trajectories by Difficulty}
\label{app:confidence_trajectories}

To complement the aggregated confidence statistics in Figure~\ref{fig:reasonorder_confidence_boxplot}, we visualize how answer token confidence evolves across diffusion steps, separated by correctness.

\begin{figure}[ht]
    \centering
    \includegraphics[width=0.95\textwidth]{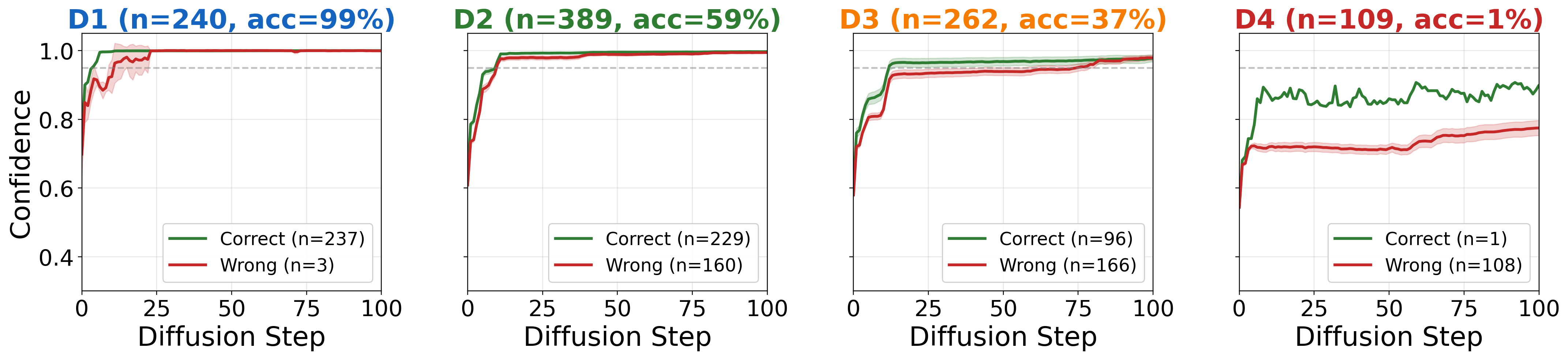}
    \caption{
    \textbf{Answer token confidence trajectories on ReasonOrderQA by difficulty level.}
    Green lines show problems with correct answers; red lines show incorrect answers. Shaded regions indicate standard deviation. 
    In D1 (acc=99\%), both correct and wrong cases rapidly converge to high confidence.
    In D2-D3, correct answers (green) achieve higher confidence earlier than wrong answers (red), suggesting that confidence trajectory shape correlates with answer correctness.
    In D4 (acc=1\%), only one problem is solved correctly; the model struggles to reach high confidence for most cases.
    }
    \label{fig:confidence_trajectories_reasonorderqa}
\end{figure}

\begin{figure}[ht]
    \centering
    \includegraphics[width=0.95\textwidth]{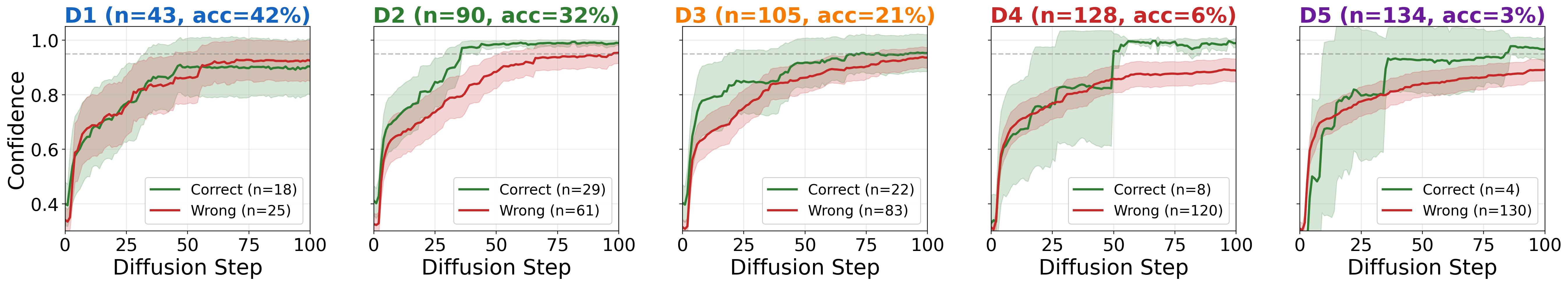}
    \caption{
    \textbf{Answer token confidence trajectories on Math500 by difficulty level (D1-D5).}
    Similar patterns emerge: correct answers (green) generally achieve higher confidence faster than wrong answers (red), though the separation is less pronounced than in ReasonOrderQA due to higher task complexity. 
    As difficulty increases, overall confidence decreases and variance increases, consistent with the boxplot results in Figure~\ref{fig:answer_confidence_boxplot}.
    }
    \label{fig:confidence_trajectories_math500}
\end{figure}

Figure~\ref{fig:confidence_trajectories_reasonorderqa} and Figure~\ref{fig:confidence_trajectories_math500} reveal that \textbf{correct answers consistently achieve higher confidence earlier} than wrong answers across both benchmarks. This pattern is most visible in intermediate difficulty levels (D2-D3 for ReasonOrderQA, D2-D4 for Math500), where the model has sufficient capacity to solve some problems but not others. The confidence trajectory shape thus serves as an implicit signal of answer correctness, supporting the mechanism identified in \S\ref{sec:mechanism}.

\section{Mathematical Formulation of Decoding Strategies}
\label{app:decoding_formulation}

To clarify the distinction between different decoding strategies used in our experiments, we provide a formal mathematical characterization of three decoding paradigms: pure autoregressive decoding, LLaDA diffusion decoding, and \texttt{left\_to\_right} decoding.

\subsection{Pure Autoregressive Decoding}

In a standard autoregressive language model, generation follows a strict left-to-right factorization:
\begin{equation}
p_{\text{AR}}(\mathbf{x}) = \prod_{i=1}^{L} p(x_i \mid x_{<i})
\end{equation}
During inference, tokens are generated sequentially:
\begin{equation}
x_i \sim p(x_i \mid x_{<i}), \quad i = 1, 2, \ldots, L
\end{equation}
At each step $i$, the model can only condition on previously generated tokens $x_{<i} = [x_1, \ldots, x_{i-1}]$. This creates a structural constraint: token $x_i$ cannot access information from future positions $x_{>i}$.

\subsection{LLaDA Diffusion Decoding}

LLaDA diffusion decoding uses a bidirectional mask predictor that estimates the entire sequence at each timestep. The generative process is:
\begin{equation}
p_\theta(\mathbf{x}_0) = \int p(\mathbf{x}_T) \prod_{t=1}^{T} p_\theta(\mathbf{x}_{t-1} \mid \mathbf{x}_t) d\mathbf{x}_{1:T}
\end{equation}
where $\mathbf{x}_T$ is a fully masked sequence, and the reverse process iteratively denoises:
\begin{equation}
\hat{\mathbf{x}}_0^{(t)} = p_\theta(\cdot \mid \mathbf{x}_t)
\end{equation}
The mask predictor $p_\theta(\cdot \mid \mathbf{x}_t)$ is a bidirectional Transformer that simultaneously predicts all masked tokens:
\begin{equation}
\hat{\mathbf{x}}_0^{(t)}[j] = p_\theta(x_j \mid \mathbf{x}_t), \quad \forall j \in \{1, \ldots, L\}
\end{equation}
where $\mathbf{x}_t$ contains both masked and unmasked tokens, and the bidirectional attention allows each position to attend to all other positions.

The default strategy for LLaDA is \texttt{low\_confidence} remasking~\citep{nie2025llada}, which selects tokens to unmask based on confidence scores:
\begin{equation}
\text{unmask}(\mathbf{x}_t, \hat{\mathbf{x}}_0^{(t)}) = \{j : \text{conf}(\hat{\mathbf{x}}_0^{(t)}[j]) > \tau_t\}, \quad \text{where } \text{conf}_j = \max_{v \in \mathcal{V}} p_\theta(x_j = v \mid \mathbf{x}_t)
\end{equation}
where $\tau_t$ is a threshold that determines how many tokens to unmask at step $t$. This allows tokens to be unmasked in any order based on their inferential certainty, not their textual position.

\subsection{Alternative Unmasking Strategies}

Beyond \texttt{low\_confidence}, we evaluate several alternative strategies that differ in how they prioritize tokens:

\begin{itemize}[leftmargin=2.5cm]
    \item \texttt{topk\_margin}~\citep{kim2025topkmargin}: Unmasks tokens based on the margin between top-1 and top-2 probabilities: $\text{margin}_j = p^{(1)}_j - p^{(2)}_j$. A larger margin indicates higher decisiveness.
    
    \item \texttt{entropy}: Unmasks tokens with lowest entropy: $H_j = -\sum_{v} p_\theta(x_j = v \mid \mathbf{x}_t) \log p_\theta(x_j = v \mid \mathbf{x}_t)$. Unlike \texttt{low\_confidence}, this considers the full distribution shape.
    
    \item \texttt{random}~\citep{austin2021structured}: Unmasks tokens uniformly at random, serving as a baseline without confidence-based selection.
\end{itemize}

\subsection{Left-to-Right Decoding} (\texttt{left\_to\_right})

The \texttt{left\_to\_right} strategy uses the same bidirectional mask predictor $p_\theta(\cdot \mid \mathbf{x}_t)$ as LLaDA diffusion, but enforces a strict left-to-right unmasking schedule. Crucially, the model still \textit{computes predictions for all tokens} at each step:
\begin{equation}
\hat{\mathbf{x}}_0^{(t)} = p_\theta(\cdot \mid \mathbf{x}_t), \quad \hat{\mathbf{x}}_0^{(t)}[j] = p_\theta(x_j \mid \mathbf{x}_t), \quad \forall j
\end{equation}
The unmasking strategy, however, is constrained to follow an autoregressive order:
\begin{equation}
\text{unmask}_{\text{AR}}(\mathbf{x}_t, \hat{\mathbf{x}}_0^{(t)}) = \{j : j \leq i_t\}
\end{equation}
where $i_t$ is the current leftmost masked position. This means:

\begin{itemize}[leftmargin=2.5cm]
    \item The model \textbf{computes} predictions for all positions (using bidirectional attention)
    \item But only \textbf{unmasks} tokens in strict left-to-right order
    \item The underlying parameterization remains diffusion-based, not AR-based
\end{itemize}

This formulation clarifies that \texttt{left\_to\_right} is not a true autoregressive model, but rather a diffusion model with an AR-like unmasking schedule. The bidirectional computation allows it to access future context internally, even though tokens are unmasked sequentially. This distinction is crucial for understanding why \texttt{left\_to\_right} may still exhibit some robustness compared to pure AR models, while being more constrained than full diffusion decoding.

\section{Prompt Templates}
\label{app:prompts}

To evaluate order robustness, we employ two distinct prompt templates that strictly enforce output formatting. The first enforces the conventional reasoning-first structure, while the second imposes an adversarial answer-first constraint.

\subsection{Standard Chain-of-Thought Prompt}

This prompt instructs the model to generate the reasoning trace prior to the final numerical result, adhering to the standard causal generation order.

\begin{promptbox}
Explain the solution with a careful chain-of-thought before giving the final numeric result, and report the final answer inside \verb|\boxed{}|.
You MUST output your answer in the following exact structure:

Reasoning:
(explain how you combined the numbers to obtain the final result)

Answer:\verb|\boxed{number}|
\end{promptbox}

\subsection{Answer-First Adversarial Prompt}

This prompt inverts the generation order, requiring the model to commit to a final answer before articulating the supporting logic.

\begin{promptbox}
Begin by stating the final numeric answer inside \verb|\boxed{}| before giving the detailed chain-of-thought explanation.
You MUST start the response with the literal text "Answer:" on its own line (no text before it).
Immediately after the heading, output the numeric answer wrapped in \verb|\boxed{}|.
Then leave a blank line and provide the reasoning under a "Reasoning:" heading.

Answer:
\verb|\boxed{number}|

Reasoning:
(explain how you combined the numbers to obtain the final result)
\end{promptbox}

\subsection{ReasonOrderQA Prompts}

For ReasonOrderQA experiments, we use specialized prompts that enforce a three-part structure: Retrieval (extracting hidden variables), Reasoning (applying arithmetic), and Answer (final result). This structure enables separate evaluation of process correctness versus outcome correctness.

\begin{promptbox}
\#\#\# YOUR TASK

You MUST output your answer in the following exact structure:

Retrieval:
(list all secret key numbers you found)

Reasoning:
(explain how you combined the numbers to obtain the final result)

Answer:
(the final num ONLY, no extra text)
\end{promptbox}

\begin{promptbox}
\#\#\# YOUR TASK

You MUST output your answer in the following exact structure:

Answer:
(the final num ONLY, no extra text)

Reasoning:
(explain how you combined the numbers to obtain the final result)

Retrieval:
(list all secret key numbers you found)
\end{promptbox}

\section{Mathematical Analysis of Confidence and Complexity}
\label{app:confidence_analysis}

We provide formal definitions of confidence, uncertainty, and difficulty to support the mechanistic analysis in \S\ref{sec:mechanism}.

\subsection{Formal Definitions}

At each diffusion step $t$, the mask predictor $p_\theta(\cdot \mid \mathbf{x}_t)$ produces a probability distribution over the vocabulary for each position $j \in \{1, \ldots, L_{\text{gen}}\}$. We define:

\begin{itemize}[leftmargin=2cm]
    \item \textbf{Confidence} for token $j$ at step $t$:
    \begin{equation}
        c_j^{(t)} = \max_{v \in \mathcal{V}} p_\theta(x_j = v \mid \mathbf{x}_t)
    \end{equation}
    where $\mathcal{V}$ is the vocabulary. This represents the model's certainty about the most likely token at position $j$.
    
    \item \textbf{Uncertainty} for token $j$ at step $t$ (measured as entropy):
    \begin{equation}
        H_j^{(t)} = -\sum_{v \in \mathcal{V}} p_\theta(x_j = v \mid \mathbf{x}_t) \log p_\theta(x_j = v \mid \mathbf{x}_t)
    \end{equation}
    Higher entropy indicates greater uncertainty about the token's identity.
    
    \item \textbf{Difficulty} for token $j$: We define difficulty as the number of diffusion steps required to achieve high confidence. A token is "difficult" if it requires more steps to converge:
    \begin{equation}
        d_j = \arg\min_t \{c_j^{(t)} > \tau\}
    \end{equation}
    where $\tau$ is a confidence threshold (e.g., 0.9). Smaller values of $d_j$ indicate tokens that converge early (easy), while larger values indicate tokens that converge late or never (hard). If a token never reaches the threshold, we set $d_j = T + 1$ to indicate maximum difficulty. This notion of difficulty is schedule-dependent and reflects convergence behavior under the specific remasking strategy used in our experiments.
\end{itemize}

\subsection{Relative Complexity Judgment}

The model's ability to judge relative complexity between tokens $i$ and $j$ depends on its capacity to distinguish their confidence distributions. We define the \textbf{confidence separation} at step $t$ as:

\begin{equation}
    \Delta_{ij}^{(t)} = |c_i^{(t)} - c_j^{(t)}|
\end{equation}

When $\Delta_{ij}^{(t)}$ is large, the model can reliably determine which token should be decoded first. When $\Delta_{ij}^{(t)}$ is small (tokens have similar confidence), the model struggles to prioritize.

The \textbf{confidence landscape variance} across all tokens at step $t$ is:

\begin{equation}
    \sigma_c^{(t)} = \sqrt{\frac{1}{L_{\text{gen}}} \sum_{j=1}^{L_{\text{gen}}} (c_j^{(t)} - \bar{c}^{(t)})^2}
\end{equation}

where $\bar{c}^{(t)} = \frac{1}{L_{\text{gen}}} \sum_{j=1}^{L_{\text{gen}}} c_j^{(t)}$ is the mean confidence. This variance measure complements the pairwise separation $\Delta_{ij}^{(t)}$: low variance indicates a flat confidence landscape where the model struggles to prioritize tokens i.e. the key factor underlying the breakdown condition identified in \S\ref{subsec:trajectory_analysis}.

\subsection{Connection to Information-Theoretic Analysis of Parallel Decoding}
\label{app:info_theory}

Our empirical findings on complexity-based ordering can be connected to recent information-theoretic analyses of parallel decoding in diffusion language models~\citep{kang2025parallelbench}.

\paragraph{Conditional Total Correlation as a Measure of Parallel Decoding Difficulty.}

\citet{kang2025parallelbench} show that the minimum achievable KL divergence for parallel decoding is lower-bounded by the conditional total correlation:

\begin{equation}
    \min_\theta D_{\text{KL}}(P_{\text{data}}(Y|X) \| P_\theta(Y|X)) \geq C(Y|X),
\end{equation}

where $C(Y|X) = \sum_{i} H(y_i|X) - H(Y|X)$ measures the total token dependencies. When $C(Y|X) > 0$, parallel decoding inevitably incurs distribution error.

\paragraph{From Total Correlation to Entropy Gap.}

While $C(Y|X)$ characterizes \textit{whether} parallel decoding will degrade quality, it does not directly explain \textit{when} order robustness holds. We propose that order robustness depends on the \textbf{entropy gap} between token types rather than overall task complexity.

For a task with answer tokens $Y_A$ and reasoning tokens $Y_R$, define the entropy gap:

\begin{equation}
    \Delta_H = \mathbb{E}_{y \in Y_A}[H(y|X, S_{<y})] - \mathbb{E}_{y \in Y_R}[H(y|X, S_{<y})],
\end{equation}

where $S_{<y}$ denotes the tokens decoded before $y$ under a given remasking trajectory. Note that $\Delta_H$ is defined with respect to the effective conditioning induced by the confidence-based remasking schedule; it characterizes the entropy gap along the actual decoding path rather than an abstract positional ordering.

\paragraph{Entropy Gap as a Proxy for Order Robustness.}

To connect entropy with our confidence-based analysis, we note that confidence can be viewed as a coarse proxy for normalized entropy. Under peaked distributions where top-1 probability mass dominates, confidence is inversely related to entropy:
\begin{equation}
    c_j^{(t)} \;\propto\; 1 - \frac{H(y_j | X, S_{<j})}{\log |\mathcal{V}|}.
\end{equation}
We use this expression only to convey monotonic intuition rather than a quantitative identity. The relationship between softmax max-probability and entropy can diverge for multi-modal or flat-tailed distributions. Nevertheless, our empirically observed confidence separation $\Delta_{ij}^{(t)}$ provides a practical proxy for the entropy gap between tokens.

This suggests a heuristic criterion for predicting order robustness:

\begin{itemize}[leftmargin=2.5cm]
    \item \textbf{Order robustness tends to hold} when $\Delta_H > \epsilon$ for some task-dependent threshold $\epsilon > 0$: answer tokens exhibit sufficiently higher entropy than reasoning tokens, leading them to be deferred by confidence-based remasking.
    
    \item \textbf{Order robustness tends to break down} when $\Delta_H \approx 0$: tokens have similar entropy, and the model cannot reliably distinguish decoding priority (our $D_2$ scenario).
\end{itemize}

\paragraph{Complementary Perspectives.}

Our analysis offers an interpretive lens that complements \citet{kang2025parallelbench}:

\begin{itemize}[leftmargin=2.5cm]
    \item \textbf{Their result:} $C(Y|X) > 0$ establishes a lower bound showing that parallel decoding cannot eliminate errors arising from token dependencies in the worst case.
    
    \item \textbf{Our interpretation:} When $\Delta_H > 0$, confidence-based remasking can exploit signals induced by these dependencies, potentially guiding the model toward correct reasoning order despite non-zero total correlation.
\end{itemize}

We emphasize that this is an empirical observation rather than a formal guarantee: the lower bound of \citet{kang2025parallelbench} does not prescribe \textit{how} diffusion samplers use dependencies, only that they cannot be ignored. Our findings suggest that token dependencies, while limiting parallel decoding in principle, can also induce uncertainty gradients that remasking-based schedules may exploit, though the precise conditions under which this occurs remain an open question.
\end{document}